\documentclass[pdflatex,sn-basic]{sn-jnl}

\jyear{2021}%

\theoremstyle{thmstyleone}%
\theoremstyle{thmstyletwo}%

\theoremstyle{thmstylethree}%

\usepackage{bm}
\usepackage{lscape}
\usepackage{xcolor}

\usepackage[section]{placeins}

\usepackage{listings}

\usepackage{booktabs}
\usepackage{multirow}
\usepackage{graphicx}
\usepackage{afterpage}

\DeclareMathOperator*{\argmin}{arg\,min} 
\newcommand{\myref}[2]{\href{#1}{\emph{\underline{#2}}}}
\newcommand{\gitlink}[2]{\href{https://github.com/#1/#2}{\emph{\underline{#2}}}}

\definecolor{light-gray}{gray}{0.95}

\newcounter{paranum}[subsubsection]
\newcommand{\subsubsubsection}{\vspace{10pt}\noindent\textbf{\thesubsubsection.\refstepcounter{paranum}\theparanum\;}\textbf}

\usepackage{caption}
\usepackage{subcaption}

\begin{document}

\title[Scientific Machine Learning through PINNs]{Scientific Machine Learning through Physics-Informed Neural Networks: Where we are and What's next}


\author[1]{\fnm{Salvatore} \sur{Cuomo}}\email{salvatore.cuomo@unina.it}

\author*[2]{\fnm{Vincenzo} \sur{Schiano Di Cola}}\email{vincenzo.schianodicola@unina.it}

\author[1]{\fnm{Fabio} \sur{Giampaolo}}\email{fabio.giampaolo@unina.it}

\author[2]{\fnm{Gianluigi} \sur{Rozza}}\email{grozza@sissa.it}

\author[3]{\fnm{Maziar} \sur{Raissi}}\email{mara4513@colorado.edu}

\author*[1]{\fnm{Francesco} \sur{Piccialli}}\email{francesco.piccialli@unina.it}

\affil*[1]{\orgdiv{Department of Mathematics and Applications "R. Caccioppoli"}, \orgname{University of Naples Federico II}, \orgaddress{\city{Napoli}, \postcode{80126}, \country{Italy} } }

\affil*[2]{\orgdiv{Department of Electrical Engineering and Information Technology}, \orgname{University of Naples Federico II}, \orgaddress{\street{Via Claudio}, \city{Napoli}, \postcode{80125}, \country{Italy} } }

\affil[2]{\orgname{SISSA – International School for Advanced Studies}, \orgaddress{\street{Street}, \city{City}, \postcode{10587}, \country{Italy}}}

\affil[3]{\orgdiv{Department of Applied Mathematics}, \orgname{University of Colorado Boulder}, \orgaddress{\street{Street}, \city{Boulder}, \postcode{610101}, \country{United States}}}


\abstract{
Physics-Informed Neural Networks (PINN) are neural networks (NNs) that encode model equations, like Partial Differential Equations (PDE), as a component of the neural network itself.
PINNs are nowadays used to solve PDEs, fractional equations, integral-differential equations, and stochastic PDEs.
This novel methodology has arisen as a multi-task learning framework in which a NN must fit observed data while reducing a PDE residual. 
This article provides a comprehensive review of the literature on PINNs: while the primary goal of the study was to characterize these networks and their related advantages and disadvantages. The review also attempts to incorporate publications on a broader range of collocation-based physics informed neural networks, which stars form the vanilla PINN, as well as many other variants, such as physics-constrained neural networks (PCNN), variational hp-VPINN, and conservative PINN (CPINN). 
The study indicates that most research has focused on customizing the PINN through different activation functions, gradient optimization techniques, neural network structures, and loss function structures.
Despite the wide range of applications for which PINNs have been used, by demonstrating their ability to be more feasible in some contexts than classical numerical techniques like Finite Element Method (FEM), advancements are still possible, most notably theoretical issues that remain unresolved.
}




\maketitle

\section{Introduction}\label{sec1}

Deep neural networks have succeeded in tasks such as computer vision, natural language processing, and game theory.
Deep Learning (DL) has transformed how categorization, pattern recognition, and regression tasks are performed across various application domains.

Deep neural networks are increasingly being used to tackle classical applied mathematics problems such as partial differential equations (PDEs) utilizing machine learning and artificial intelligence approaches.
Due to, for example, significant nonlinearities, convection dominance, or shocks, some PDEs are notoriously difficult to solve using standard numerical approaches. Deep learning has recently emerged as a new paradigm of scientific computing 
thanks to neural networks' universal approximation and great expressivity.
Recent studies have shown deep learning to be a promising method for building meta-models for fast predictions of dynamic systems. In particular, NNs have proven to represent the underlying nonlinear input-output relationship in complex systems. %
Unfortunately, dealing with such high dimensional-complex systems are not exempt from the curse of dimensionality, which Bellman first described in the context of optimal control problems \citep{bellman1966dynamic}.
%
However, machine learning-based algorithms are promising for solving PDEs \citep{Ble2021_ThreeWaysSolve_ErnBE}.

Indeed, \cite{Ble2021_ThreeWaysSolve_ErnBE} consider machine learning-based PDE solution approaches will continue to be an important study subject in the next years as deep learning develops in methodological, theoretical, and algorithmic developments. 
Simple neural network models, such as MLPs with few hidden layers, were used in early work for solving differential equations \citep{Lag1998_ArtificialNeuralNetworks_LikLLF}.
Modern methods, based on NN techniques, take advantage of optimization frameworks and auto-differentiation, like \cite{Ber2018_UnifiedDeepArtificial_NysBN} that suggested a unified deep neural network technique for estimating PDE solutions.
Furthermore, it is envisioned that DNN will be able to create an interpretable hybrid Earth system model based on neural networks for Earth and climate sciences \citep{Irr2021_TowardsNeuralEarth_BoeIBS}.


Nowadays, the literature does not have a clear nomenclature for integrating previous knowledge of a physical phenomenon with deep learning.
`Physics-informed,' `physics-based,' `physics-guided,' and `theory-guided' are often some used terms. \cite{Kim2021_KnowledgeIntegrationDeep_KimKKLL}  developed the overall taxonomy of what they called informed deep learning, followed by a literature review in the field of dynamical systems.
Their taxonomy is organized into three conceptual stages: (i) what kind of deep neural network is used, (ii) how physical knowledge is represented, and (iii) how physical information is integrated. 
Inspired by their work, we will investigate PINNs, a 2017 framework, and demonstrate how neural network features are used, how physical information is supplied, and what physical problems have been solved in the literature. 

\subsection{What the PINNs are} 
%
Physics-Informed Neural Networks (PINNs) are a scientific machine learning technique used to solve problems involving Partial Differential Equations (PDEs).
PINNs approximate PDE solutions by training a neural network to minimize a loss function; it includes terms reflecting the initial and boundary conditions along the space-time domain's boundary and the PDE residual at selected points in the domain (called collocation point). 
PINNs are deep-learning networks that, given an input point in the integration domain, produce an estimated solution in that point of a differential equation after training.
Incorporating a residual network that encodes the governing physics equations is a significant novelty with PINNs.
The basic concept behind PINN training is that it can be thought of as an unsupervised strategy that does not require labelled data, such as results from prior simulations or experiments.
The PINN algorithm is essentially a mesh-free technique that finds PDE solutions by converting the problem of directly solving the governing equations into a loss function optimization problem.
It works by integrating the mathematical model into the network and reinforcing the loss function with a residual term from the governing equation, which acts as a penalizing term to restrict the space of acceptable solutions.


PINNs take into account the underlying PDE, i.e. the physics of the problem, rather than attempting to deduce the solution based solely on data, i.e. by fitting a neural network to a set of state-value pairs.
The idea of creating physics-informed learning machines that employ systematically structured prior knowledge about the solution can be traced back to earlier research by \cite{Owh2015_BayesianNumericalHomogenization_Owh}, which revealed the promising technique of leveraging such prior knowledge.
\cite{Rai2017_InferringSolutionsDifferential_PerRPK, Rai2017_MachineLearningLinear_PerRPK} used Gaussian process regression to construct representations of linear operator functionals, accurately inferring the solution and providing uncertainty estimates for a variety of physical problems; this was then extended in \citep{Rai2018_HiddenPhysicsModels_KarRK, Rai2018_NumericalGaussianProcesses_PerRPK}.
%
PINNs were introduced in 2017 as a new class of data-driven solvers in a two-part article \citep{Rai2017_PhysicsInformedDeep1_PerRPK,Rai2017_PhysicsInformedDeep2_PerRPK} published in a merged version afterwards in 2019 \citep{Rai2019_PhysicsInformedNeural_PerRPK}.
\cite{Rai2019_PhysicsInformedNeural_PerRPK}
 introduce and illustrate the PINN approach for solving nonlinear PDEs, like Schrödinger, Burgers, and Allen-Cahn equations. 
They created physics-informed neural networks (PINNs) 
which can handle both forward problems of estimating the solutions of governing mathematical models and inverse problems, where the model parameters are learnt from observable data.

The concept of incorporating prior knowledge into a machine learning algorithm is not entirely novel.
In fact \cite{Dis1994_NeuralNetworkBased_PhaDP} can be considered one of the first PINNs. 
This paper followed the results of the universal approximation achievements of the late 1980s, \citep{Hor1989_MultilayerFeedforwardNetworks_StiHSW}; then in the early 90s several methodologies were proposed to use neural networks to approximate PDEs, like the work on constrained neural networks
\citep{Psi1992_HybridNeuralNetwork_UngPU, Lag1998_ArtificialNeuralNetworks_LikLLF}
or 
\citep{Lee1990_NeuralAlgorithmSolving_KanLK}.
In particular \cite{Dis1994_NeuralNetworkBased_PhaDP}
employed a simple neural networks to approximate a PDE, where the neural network's output was a single value that was designed to approximate the solution value at the specified input position. The network had two hidden layers, with 3, 5 or 10 nodes for each layer.
The network loss function approximated the $L^2$ error of the approximation on the interior and boundary of the domain using point-collocation.
While, the loss is evaluated using a quasi-Newtonian approach and the gradients are evaluated using finite difference.
\\
In \cite{Lag1998_ArtificialNeuralNetworks_LikLLF}, the solution of a differential equation is expressed as a constant term and an adjustable term with unknown parameters, the best parameter values are determined via a neural network. However, their method only works for problems with regular borders.
\cite{Lag2000_NeuralNetworkMethods_LikLLP} extends the method to problems with irregular borders.

As computing power increased during the 2000s, increasingly sophisticated models with more parameters and numerous layers became the standard \cite{Oez2021_PoissonCnnConvolutional_HamOeHL}. 
Different deep model using MLP, were introduced, also using Radial Basis Function \cite{Kum2011_MultilayerPerceptronsRadial_YadKY}.

Research into the use of NNs to solve PDEs continued to gain traction in the late 2010s, thanks to advancements in the hardware used to run NN training, the discovery of better practices for training NNs, and the availability of open-source packages, such as Tensorflow \citep{Hag2021_SciannKerastensorflowWrapper_JuaHJ}, and the availability of Automatic differentiation in such packages \citep{Pas2017_AutomaticDifferentiationPytorch_GroPGC}.

Finally, more recent advancements by 
\cite{Kon2018_GeneralizationEquivarianceConvolution_TriKT}, 
and 
\cite{Mal2016_UnderstandingDeepConvolutional_Mal}, 
brought to 
\cite{Rai2019_PhysicsInformedNeural_PerRPK} solution that extended previous notions while also introducing fundamentally new approaches, such as a discrete time-stepping scheme that efficiently leverages the predictive ability of neural networks  \citep{Kol2021_PhysicsInformedNeural_DAKDJH}.
The fact that the framework could be utilized directly by plugging it into any differential problem simplified the learning curve for beginning to use such, and it was the first step for many researchers who wanted to solve  their problems with a Neural network approach \citep{Mar2021_OldNewCan_Mar}. 
The success of the PINNs can be seen from the rate at which \cite{Rai2019_PhysicsInformedNeural_PerRPK} is cited, and the exponentially growing number of citations in the recent years (Figure~\ref{fig:time}).

However, PINN is not the only NN framework utilized to solve PDEs.
Various frameworks have been proposed in recent years,  and, while not exhaustive, we have attempted to highlight the most relevant ones in this paragraph. 

The Deep Ritz method (DRM) \citep{Wei2018_DeepRitzMethod_YuWY}, where the loss is defined as the energy of the problem's solution.

Then there approaches based on the Galerkin method, or Petrov–Galerkin method, where the loss is given by multiplying the residual by a test function, and when is the volumetric residual we have a Deep Galerkin Method (DGM) \citep{Sir2018_DgmDeepLearning_SpiSS}.
Whereas, when a Galerkin approach is used on collocation points the framework is a variant of PINNs, i.e. a hp-VPINN \cite{Kha2021_HpVpinnsVariational_ZhaKZK}.


Within the a collocation based approach, i.e. PINN methodology  \citep{Rai2019_PhysicsInformedNeural_PerRPK, Yan2019_AdversarialUncertaintyQuantification_PerYP, Men2020_PpinnPararealPhysics_LiMLZK}, many other variants were proposed, as  the variational hp-VPINN, as well as conservative PINN (CPINN)\citep{Jag2020_ConservativePhysicsInformed_KhaJKK}.
Another approach is  physics-constrained neural networks (PCNNs) \citep{Zhu2019_PhysicsConstrainedDeep_ZabZZKP, Sun2020_SurrogateModelingFluid_GaoSGPW, Liu2021_DualDimerMethod_WanLW}.
while PINNs incorporate both the PDE and its initial/boundary conditions (soft BC) in the training loss function, PCNNs, are ``data-free'' NNs, i.e. they enforce the initial/boundary conditions (hard BC) via a custom NN architecture while embedding the PDE in the training loss. 
This soft form technique is described in  \cite{Rai2019_PhysicsInformedNeural_PerRPK}, where the term ``physics-informed neural networks'' was coined (PINNs).

Because there are more articles on PINN than any other specific variant, such as PCNN, hp-VPINN, CPINN, and so on, this review will primarily focus on PINN, with some discussion of the various variants that have emerged, that is, NN architectures that solve differential equations based on collocation points. 

Finally, the acronym PINN will be written in its singular form rather than its plural form in this review, as it is considered representative of a family of neural networks of the same type.

Various review papers involving PINN have been published.
About the potentials, limitations and applications for forward and inverse problems  \citep{Kar2021_PhysicsInformedMachine_KevKKL}
for three-dimensional flows \citep{Cai2021_PhysicsInformedNeural_MaoCMW},
or a comparison with other ML techniques \citep{Ble2021_ThreeWaysSolve_ErnBE}. 
%
An introductory course on PINNs that covers the fundamentals of Machine Learning and Neural Networks can be found from \cite{Kol2021_PhysicsInformedNeural_DAKDJH}.
PINN is also compared against other methods that can be applied to solve PDEs, like the one based on the Feynman–Kac theorem \citep{Ble2021_ThreeWaysSolve_ErnBE}.
Finally, PINNs have also been extended to solve integrodifferential equations (IDEs)\citep{Pan2019_FpinnsFractionalPhysics_LuPLK, Yua2022_PinnAuxiliaryPhysics_NiYNDH} or stochastic differential equations (SDEs) \citep{Yan2020_PhysicsInformedGenerative_ZhaYZK}. 

%
%
%
%

Being able to learn PDEs, PINNs have several advantages over conventional methods.
PINNs, in particular, are mesh-free methods that enable on-demand solution computation after a training stage, and they allow solutions to be made differentiable using analytical gradients.
Finally, they provide an easy way to solve forward jointly and inverse problems using the same optimization problem.
In addition to solving differential equations (the forward problem), PINNs may be used to solve inverse problems such as 
characterizing fluid flows from sensor data. 
In fact, the same code used to solve forward problems can be used to solve inverse problems with minimal modification. 
Indeed, PINNs can address PDEs in domains with very complicated geometries or in very high dimensions that are all difficult to numerically simulate as well as inverse problems and constrained optimization problems.
%


\subsection{What is this review about} 
In this survey, we focus on how PINNs are used to address different scientific computing problems, the building blocks of PINNs, the aspects related to learning theory, what toolsets are already available, future directions and recent trends, and issues regarding accuracy and convergence.
According to different problems, we show how PINNs solvers have been customized in literature by configuring the depth, layer size, activation functions and using transfer learning.

This article can be considered as an extensive literature review on PINNs.
It was mostly conducted by searching Scopus for the terms:
\\
\noindent
\texttt{((physic* OR physical) W/2 (informed OR constrained) W/2 ``neural network’’)
}
\\
\noindent
%
%
The primary research question was to determine what PINNs are and their associated benefits and drawbacks.
The research also focused on the outputs from the CRUNCH research group in the Division of Applied Mathematics at Brown University and then on the (Physics-Informed Learning Machines for Multiscale and Multiphysics Problems) PhILMs Center, which is a collaboration with the Pacific Northwest National Laboratory.
In such a frame, the primary authors who have been involved in this literature research are Karniadakis G.E., Perdikaris P., and Raissi M.
Additionally, the review considered studies that addressed a broader topic than PINNs, namely physics-guided neural networks, as well as physics-informed machine learning and deep learning.

\begin{figure}[htbp]
    \centering
    \includegraphics[width=0.99\textwidth]{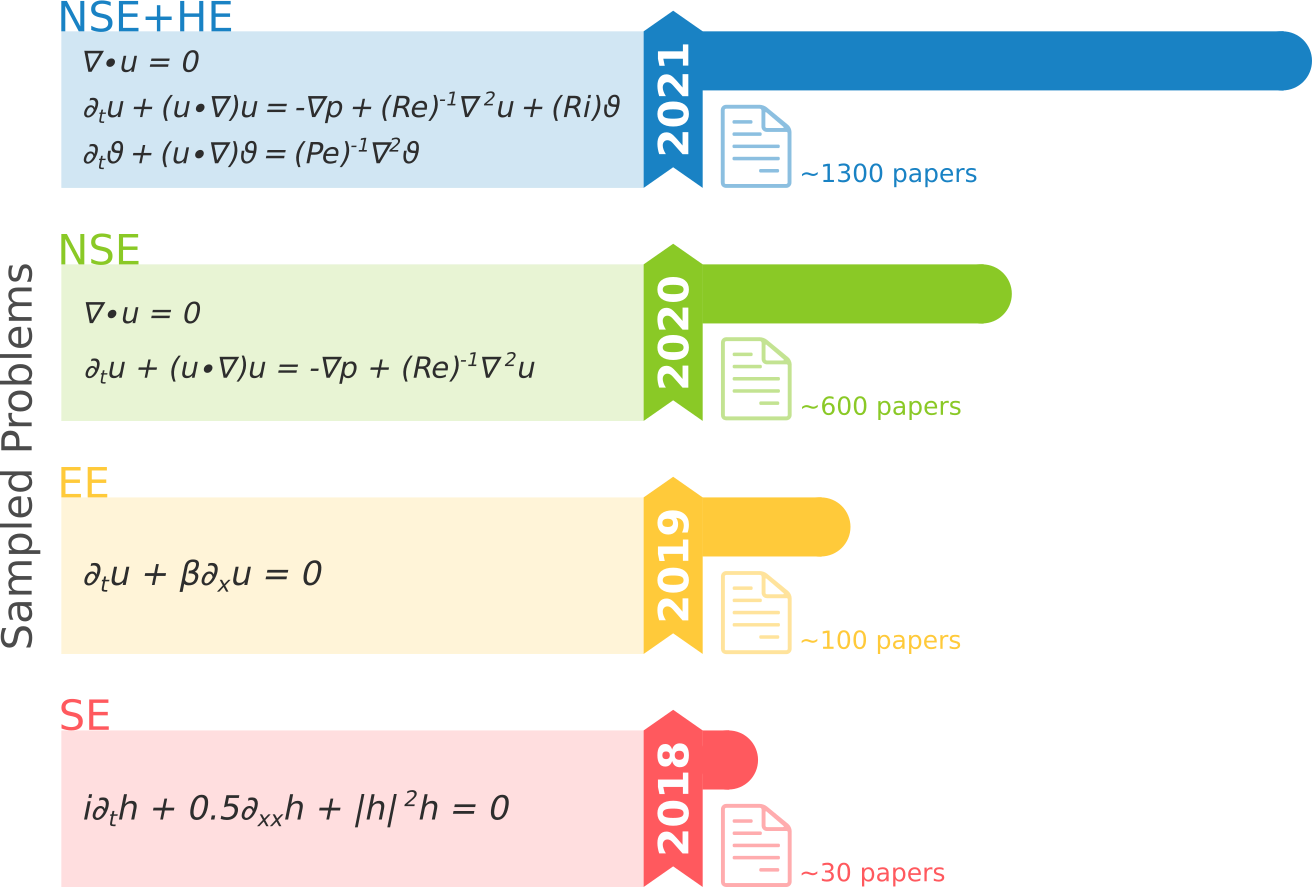}
    \caption{%
A number of papers related to PINNs (on the right) addressed problems on which PINNs are applied (on the left) by year.
PINN is having a significant impact in the literature and in scientific applications.
The number of papers referencing Raissi or including PINN in their abstract title or keywords is increasing exponentially.
The number of papers about PINN more than quintupled between 2019 and 2020, and there were twice as many new papers published in 2021 as there were in 2020. 
Since Raissi's first papers on arXiv in 2019 \citep{Rai2019_PhysicsInformedNeural_PerRPK}, a boost in citations can be seen in late 2018 and 2019.
On the left, we display a sample problem solved in that year by one of the articles, specifically a type of time-dependent equation. 
Some of the addressed problems to be solved in the first vanilla PINN, for example, were the Allen–Cahn equation, the Korteweg–de Vries equation, or the 1D nonlinear Shr\"{o}dinger problem (SE). 
By the end of 2019 \cite{Mao2020_PhysicsInformedNeural_JagMJK} solved with the same PINN Euler equations (EE) that model high-speed aerodynamic flows.
By the end of 2020 \cite{Jin2021_NsfnetsNavierStokes_CaiJCLK} solved the incompressible Navier-Stokes equations (NSE).
Finally, in 2021 \cite{Cai2021_PhysicsInformedNeural_WanCWW} coupled the Navier–Stokes equations with the
corresponding temperature equation for analyzing heat flow convection (NSE+HE). 
}
    \label{fig:time}
\end{figure}

Figure~\ref{fig:time} summarizes what the influence of PINN is in today's literature and applications.

%


\section{The building blocks of a PINN}\label{sec2}


Physically-informed neural networks can address problems that are described by few data, or noisy experiment observations.
Because they can use known data while adhering to any given physical law specified by general nonlinear partial differential equations, PINNs can also be considered neural networks that deal with supervised learning problems
\citep{Gos2020_TransferLearningEnhanced_AniGACR}.
PINNs can solve differential equations expressed, in the most general form, like:
\begin{equation}\label{eq:general_form}
  \begin{aligned}
    \mathcal{F}(\bm{u}(\bm{z}); \gamma ) &= \bm{f}(\bm{z}) \quad &&  \bm{z} \text{ in } \Omega,\\
    \mathcal{B}(\bm{u}(\bm{z})) &= \bm{g}(\bm{z})  \quad &&  \bm{z} \text{ in } \partial\Omega
  \end{aligned}
\end{equation} 
defined on the domain $\Omega \subset \mathbb{R}^d$ with the boundary $\partial \Omega$.
Where 
$\bm{z} := [x_1,\ldots,x_{d-1};t]$ indicates the space-time coordinate vector, 
$\bm{u}$ represents the unknown solution, $\gamma$ are the parameters related to the physics, $\bm{f}$ is the function identifying the data of the problem and $\mathcal{F}$ is the non linear differential operator.
Finally, since the initial condition can actually be considered as a type of Dirichlet boundary condition on the spatio-temporal domain, it is possible to 
denote 
$\mathcal{B}$ as the operator indicating arbitrary initial or boundary conditions related to the problem and $\bm{g}$ the boundary function. Indeed, the boundary conditions can be Dirichlet, Neumann, Robin, or periodic boundary conditions.

Equation~\eqref{eq:general_form} 
can describe numerous physical systems including 
both forward and inverse problems.
The goal of forward problems is to find the function $\bm{u}$ for every $\bm{z}$, where $\gamma$ are specified parameters.
In the case of the inverse problem, $\gamma$ must also be determined from the data.
The reader can find an operator based mathematical formulation
of equation~\eqref{eq:general_form}
in the work of  \cite{Mis2021_EstimatesGeneralizationError_MolMM}.
In the PINN methodology, $\bm{u}(\bm{z})$ is computationally predicted by a NN, parametrized  by a set of parameters $\theta$, giving rise to an approximation $$\hat{\bm{u}}_\theta(\bm{z}) \approx \bm{u}(\bm{z});$$
where $\hat{(\cdot)}_\theta$ denotes a NN approximation realized with $\theta$.

In this context, where forward and inverse problems are analyzed in the same framework, and given that PINN can adequately solve both problems, we will use $\theta$ to represent both the vector of all unknown parameters in the neural network that represents the surrogate model and unknown parameters $\gamma$ in the case of an inverse problem.


In such a context, the NN must learn to approximate the differential equations through finding $\theta$ that define the NN by minimizing a loss function that depends on the differential equation $\mathcal{L}_\mathcal{F}$, the boundary conditions $ \mathcal{L}_\mathcal{B}$, and eventually some known data
$\mathcal{L}_{data}$, each of them adequately weighted:
\begin{equation}\label{eq:general_loss}
\theta^* =  \mathop{\argmin}_{\theta} \left(
\omega_\mathcal{F} \mathcal{L}_\mathcal{F}(\theta) +
\omega_\mathcal{B} \mathcal{L}_\mathcal{B}(\theta) +
\omega_{d} \mathcal{L}_{data} (\theta) 
\right).
\end{equation}
To summarize, PINN can be viewed as an unsupervised learning approach when they are trained solely using physical equations and boundary conditions for forward problems; however, for inverse problems or when some physical properties are derived from data that may be noisy, PINN can be considered supervised learning methodologies.

In the following paragraph, we will discuss the types of NN used to approximate  $\bm{u}(\bm{z})$, how the information derived by $\mathcal{F}$ is incorporated in the model, and how the NN learns from the equations and additional given data.

\begin{figure}[htbp]
    \centering
    \includegraphics[width=0.99\textwidth]{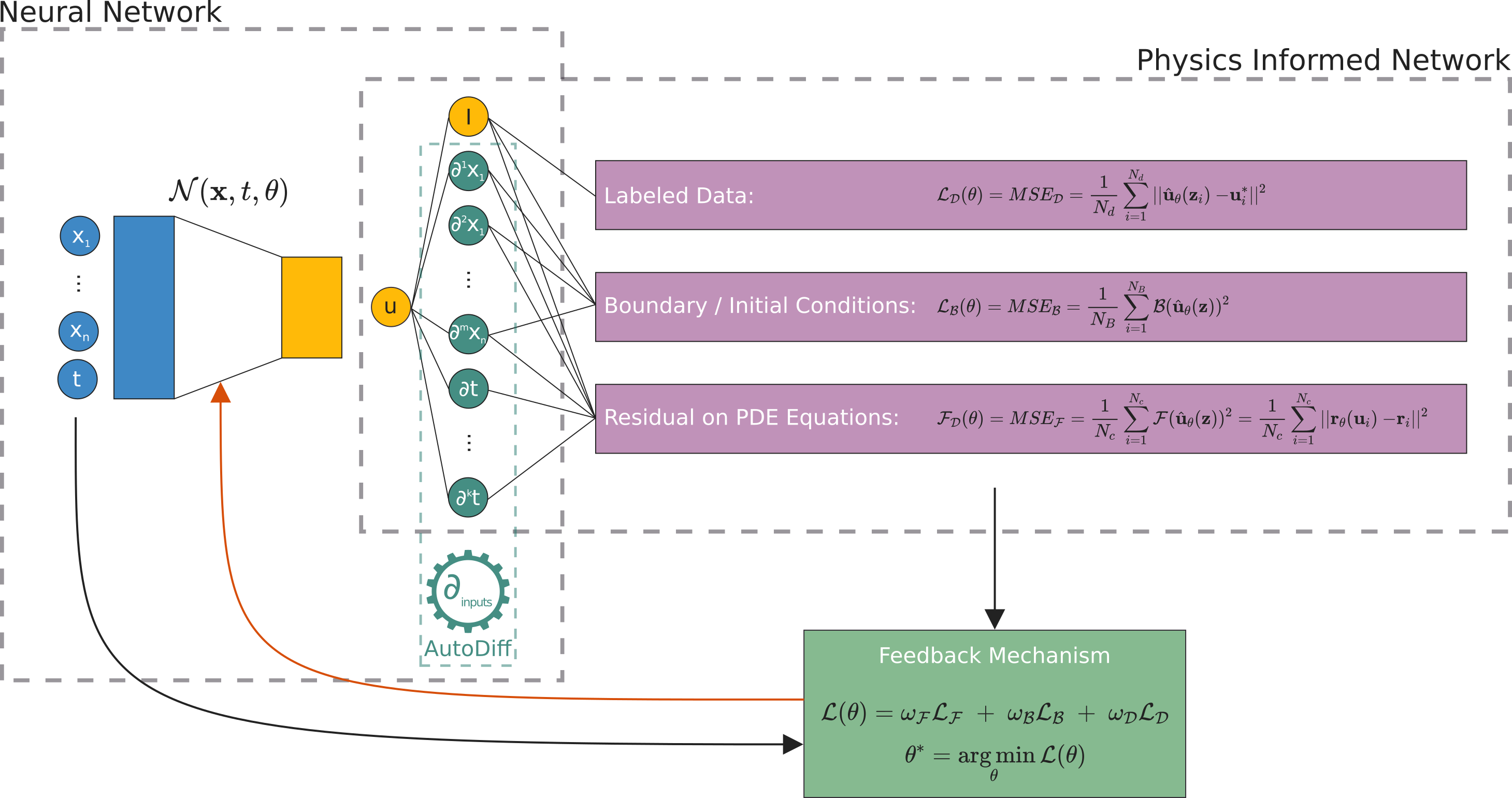}
    \caption{%
Physics-informed neural networks building blocks. 
PINNs are made up of differential equation residual (loss) terms, as well as initial  and boundary conditions.
The network's inputs (variables) are transformed into network outputs (the field $\bm{u}$). 
The network is defined by $\theta$.
The second area is the physics informed network, which takes the output field $\bm{u}$ and computes the derivative using the given equations. The boundary/initial condition is also evaluated (if it has not already been hard-encoded in the neural network), and also the labeled data observations are calculated (in case there is any data available).
The final step with all of these residuals is the feedback mechanism, that minimizes the loss, using an optimizer, according to some learning rate in order to obtain the NN parameters  $\theta$. 
}
    \label{fig:PINN}
\end{figure}

Figure~\ref{fig:PINN} summarizes all the PINN's building blocks discussed in this section.
PINN are composed of three components: a neural network, a physics-informed network, and a feedback mechanism.
The first block is a NN, $\hat{\bm{u}}_\theta$,  that accepts vector variables $\bm{z}$  from the equation~\eqref{eq:general_form} and outputs the filed value $\bm{u}$.
The second block can be thought of PINN's functional component, as it computes the derivative to determine the losses of equation terms, as well as the terms of the initial and boundary conditions of equation~\eqref{eq:general_loss}.
Generally, the first two blocks are linked using algorithmic differentiation, which is used to inject physical equations into the NN during the training phase.
Thus, the feedback mechanism minimizes the loss according to some learning rate, in order to fix the NN parameters vector $\theta$ of the NN $\hat{\bm{u}}_\theta$. 
In the following, it will be clear from the context to what network we are referring to, whether the NN or the functional network that derives the physical information.

\subsection{Neural Network Architecture}

The representational ability of neural networks is well established.
According to the universal approximation theorem, any continuous function can be arbitrarily closely approximated by a multi-layer perceptron with only one hidden layer and a finite number of neurons \citep{Hor1989_MultilayerFeedforwardNetworks_StiHSW, Cyb1989_ApproximationSuperpositionsSigmoidal_Cyb, Yar2017_ErrorBoundsApproximations_Yar, Ber2019_SurveyDeepLearning_BucBBCC}.
While neural networks can express very complex functions compactly, determining the precise parameters (weights and biases) required to solve a specific PDE can be difficult \citep{Wah2021_PinneikEikonalSolution_HagWHA}.
%
%
Furthermore, identifying the appropriate artificial neural network (ANN) architecture can be challenging.
There are two approaches: shallow learning networks, which have a single hidden layer and can still produce any non-linear continuous function, and deep neural networks (DNN), a type of ANN that uses more than two layers of neural networks to model complicated relationships
\citep{Ald2020_DeepLearningApproaches_DerADE}.
Finally, the taxonomy of various Deep Learning architectures is still a matter of research  \citep{Muh2021_DeepLearningApplication_AseMAC, Ald2020_DeepLearningApproaches_DerADE, Sen2020_ReviewDeepLearning_BasSBS}

The main architectures covered in the Deep Learning  literature include fully connected feed-forward networks (called FFNN, or also FNN, FCNN, or FF-DNN), convolutional neural networks (CNNs), and recurrent neural networks (RNNs) \citep{LeC2015_DeepLearning_BenLBH, Che2018_RiseDeepLearning_EngCEW}.
Other more recent architectures encompass Auto-Encoder (AE), Deep Belief Network (DBN), Generative Adversarial Network (GAN), and Bayesian Deep Learning (BDL) \citep{Alo2019_StateArtSurvey_TahATY, Ber2019_SurveyDeepLearning_BucBBCC}.%

Various PINN extensions have been investigated, based on some of these networks. An example is estimating the PINN solution's uncertainty using Bayesian neural networks. Alternatively, when training CNNs and RNNs, finite-difference filters can be used to construct the differential equation residual in a PINN-like loss function. 
This section will illustrate all of these examples and more; first, we will define the mathematical framework for characterizing the various networks.

According to 
\cite{Cat2018_GenericRepresentationNeural_ChaCC}
a generic deep neural network with $L$ layers, can be represented  as the composition of $L$ functions $f_i(x_i,\theta_i)$  where $x_i$ are known as state variables, and
 $\theta_i$ denote the set of parameters for the $i$-th layer. So a function $\bm{u}(x)$ is approximated as
\begin{equation}
\label{eq:dnn}
\bm{u}_{\theta}(x) = f_L \circ f_{L-1}  \ldots   \circ f_1(x).
\end{equation} 
where
each $f_i$ is defined on two inner product spaces $E_i$ and $H_i$, being $f_i\in E_i \times H_i$
and 
the layer composition, represented  by $\circ$, 
is to be read as $f_2 \circ f_1 (x) = f_2(f_1(x))$.


Since \cite{Rai2019_PhysicsInformedNeural_PerRPK} original vanilla PINN, the majority of solutions have used feed-forward neural networks. 
However, some researchers have experimented with different types of neural networks to see how they affect overall PINN performance, in particular with CNN, RNN, and GAN, and there seems there has been not yet a full investigation of other networks within the PINN framework.
In this subsection, we attempt to synthesize the mainstream solutions, utilizing feed-forward networks, and all of the other versions. 
Table~\ref{tab:NeuralNetworks} contains a synthesis reporting the kind of neural network and the paper that implemented it.

\subsubsection{Feed-forward neural network}
A feed-forward neural network, also known as a multi-layer perceptron (MLP), is a collection of neurons organized in layers that perform calculations sequentially through the layers. 
Feed-forward networks, also known as multilayer perceptrons (MLP), are DNNs with several hidden layers that only move forward (no loopback). 
In a fully connected neural network, neurons in adjacent layers are connected, whereas neurons inside a single layer are not linked. Each neuron is a mathematical operation that applies an activation function to the weighted sum of it's own inputs plus a bias factor \citep{Wah2021_PinneikEikonalSolution_HagWHA}.
Given an input $x \in \Omega$ an MLP transforms it to an output, through a layer of units (neurons) which compose of affine-linear maps between units (in successive layers) and scalar non-linear activation functions within units, resulting in a composition of functions.
So for MLP, it is possible to specify~\eqref{eq:dnn}  as
$$
f_i (x_i ; W_i, b_i) = \alpha_i (W_i \cdot x_i + b_i)
$$
equipping each $E_i$ and $H_i$ with the standard Euclidean inner product, i.e. $E=H=\mathbb{R}$ \citep{Cat2018_SpecificNetworkDescriptions_ChaCC}, and $\alpha_i$ is a scalar (non-linear) activation function.  The machine learning literature has studied several different activation functions, which we shall discuss later in this section.
Equation~\eqref{eq:dnn} can also be rewritten in conformity with the notation used in \cite{Mis2021_EstimatesGeneralizationError_MolMM}:
\begin{equation}
\label{eq:ann}
\bm{u}_{\theta}(x) = C_K \circ\alpha \circ C_{K-1}  \ldots  \circ \alpha \circ C_1(x),
\end{equation} 
where for any $1 \leq k \leq K$, it it is defined
\begin{equation}
\label{eq:C}
C_k (x_k) = W_k x_k + b_k.
\end{equation}
Thus a neural network consists of an input layer, an output layer, and $(K-2)$ hidden layers.

\paragraph{FFNN architectures}
While the ideal DNN architecture is still an ongoing research; papers implementing PINN have attempted to empirically optimise the architecture's characteristics, such as the number of layers and neurons in each layer.
Smaller DNNs may be unable to effectively approximate unknown functions, whereas too large DNNs may be difficult to train, particularly with small datasets. 
%
\cite{Rai2019_PhysicsInformedNeural_PerRPK} used different typologies of DNN, for each problem, like a 5-layer deep neural network with 100 neurons per layer, an DNN with 4 hidden layers and 200 neurons per layer or a  9 layers with 20 neuron each layer.
\cite{Tar2020_PhysicsInformedDeep_MarTMP} 
empirically determine the feedforward network size, in particular they use three hidden layers and 50 units per layer, all with an hyperbolic tangent activation function.
Another example of how the differential problem  affects network architecture can be found in \cite{Kha2021_HpVpinnsVariational_ZhaKZK} for their hp-VPINN. The architecture is implemented with four layers and twenty neurons per layer, but for an advection equation with a double discontinuity of the exact solution, they use an eight-layered deep network. 
For a constrained approach,  by utilizing a specific portion of the NN to satisfy the required boundary conditions, \cite{Zhu2021_MachineLearningMetal_LiuZLY} use five hidden layers and 250 neurons per layer to constitute the fully connected neural network.
Bringing the number of layers higher, 
in PINNeik \citep{Wah2021_PinneikEikonalSolution_HagWHA}, a DNN with ten hidden layers containing twenty neurons each is utilized, with a locally adaptive inverse tangent function as the activation function for all hidden layers except the final layer, which has a linear activation function. 
%
\cite{He2020_PhysicsInformedNeural_BarHBTT} examines the effect of neural network size on state estimation accuracy.
They begin by experimenting with various hidden layer sizes ranging from three to five, while maintaining a value of 32 neurons per layer.
Then they set the number of hidden layers to three, the activation function to  hyperbolic tangent, while varying the number of neurons in each hidden layer. 
Other publications have attempted to understand how the number of layers, neurons, and activation functions effect the NN's approximation quality with respect to the problem to be solved, like \citep{Ble2021_ThreeWaysSolve_ErnBE}.  



\paragraph{Multiple FFNN}  
Although many publications employ a single fully connected network, a rising number of research papers have been addressing PINN with multiple fully connected network blended together, e.g. to approximate 
specific equation of a larger mathematical model.
An architecture composed of five the feed-forward neural network is proposed by \cite{Hag2021_PhysicsInformedDeep_RaiHRM}.
In a two-phase Stefan problem, discussed later in this review and in \cite{Cai2021_PhysicsInformedNeural_WanCWW},
a DNN is used to model the unknown interface between two different material phases, while another DNN describes the two temperature distributions of the phases.
Instead of a single NN across the entire domain, \cite{Mos2021_FiniteBasisPhysics_MarMMN} suggests using multiple neural networks, one for each subdomain.
Finally, \cite{Lu2021_LearningNonlinearOperators_JinLJP} employed a pair of DNN, one for encoding the input space (branch net) and the other for encoding the domain of the output functions (trunk net).
This architecture, known as DeepONet, is particularly generic because no requirements are made to the topology of the branch or trunk network, despite the fact that the two sub-networks have been implemented as FFNNs as in \cite{Lin2021_SeamlessMultiscaleOperator_MaxLMLK}.

\paragraph{Shallow networks}
To overcome some difficulties,
various researchers have also tried to investigate shallower network solutions: these can be sparse neural networks, instead of fully connected architectures, or more likely single hidden layers as ELM (Extreme Learning Machine) \citep{Hua2011_ExtremeLearningMachines_WanHWL}.
When compared to the shallow architecture, more hidden layers aid in the modeling of complicated nonlinear relationships \citep{Sen2020_ReviewDeepLearning_BasSBS}, however, 
using PINNs for real problems can result in deep networks with many layers associated with high training costs and efficiency issues. For this reason, not only deep neural networks have been employed for PINNs but also shallow ANN are reported in the literature. 
%
X-TFC, developed by \cite{Sch2021_ExtremeTheoryFunctional_FurSFL}, employs a single-layer NN trained using the ELM algorithm. 
While PIELM \citep{Dwi2020_PhysicsInformedExtreme_SriDS} is proposed as a faster alternative, using a hybrid neural network-based method that combines two ideas from PINN and ELM. 
ELM only updates the weights of the outer layer, leaving the weights of the inner layer unchanged.
\\
\noindent
Finally, in \cite{Ram2021_SpinnSparsePhysics_RamRR}
a Sparse, Physics-based, and partially Interpretable Neural Networks (SPINN) is proposed.
The authors suggest a sparse architecture, using kernel networks, that yields interpretable results while requiring fewer parameters than fully connected solutions.
They consider various kernels such as Radial Basis Functions (RBF), softplus hat, or Gaussian kernels, and apply their proof of concept architecture to a variety of mathematical problems.

\paragraph{Activation function}
The activation function has a significant impact on DNN training performance.
ReLU, Sigmoid, Tanh are commonly used activations \citep{Sun2020_SurrogateModelingFluid_GaoSGPW}.
Recent research has recommended training an adjustable activation function like  Swish, which is defined as $x\cdot \textrm{Sigmoid}(\beta x)$ and $\beta$ is a trainable parameter, 
and where $\textrm{Sigmoid}$ is supposed to be a general sigmoid curve, an S-shaped function, or in some cases a logistic function. 
Among the most used activation functions there are logistic sigmoid, hyperbolic tangent, ReLu, and leaky ReLu.
Most authors tend to use the infinitely differentiable hyperbolic tangent activation function $\alpha(x) = \tanh(x)$ \citep{He2020_PhysicsInformedNeural_BarHBTT},
whereas \cite{Che2021_DeepLearningMethod_ZhaCZ} use a Resnet block to improve the stability of the fully-connected neural network (FC-NN).
They also prove that Swish activation function outperforms the others in terms of enhancing the neural network's convergence rate and accuracy. 
Nonetheless, because of the second order derivative evaluation, it is pivotal to choose the activation function in a PINN framework with caution. 
For example, while rescaling the PDE to dimensionless form, it is preferable to choose a range of $[0,1]$ rather than a wider domain, because most activation functions (such as Sigmoid, Tanh, Swish) are nonlinear near $0$. 
Moreover the regularity of PINNs can be ensured by using smooth activation functions like as the sigmoid and hyperbolic tangent, allowing estimations of PINN generalization error to hold true \citep{Mis2021_EstimatesGeneralizationError_MolMM}. 

\subsubsection{Convolutional neural networks}

Convolutional neural networks (ConvNet or CNN) are intended to process data in the form of several arrays, for example a color image made of three 2D arrays. 
CNNs usually have a number of convolution and pooling layers. 
The convolution layer is made up of a collection of filters (or kernels) that convolve across the full input  rather than general matrix multiplication. 
The pooling layer instead performs subsampling, reducing the dimensionality. 

For CNNs, according to \cite{Cat2018_SpecificNetworkDescriptions_ChaCC}, the layerwise function $f$ can be written as
$$
f_i (x_i; W_i) = \Phi_i (\alpha_i(\mathcal{C}_i(W_i,x_i)))
$$
where $\alpha$ is an elementwise nonlinearity,  $\Phi$ is the max-pooling map, and $\mathcal{C}$  the convolution operator.

It is worth noting that the convolution operation preserves translations 
and pooling is unaffected by minor data translations. 
This is applied to input image properties, such as corners, edges, and so on, that are translationally invariant, and will still be represented in the convolution layer's output.

As a result, CNNs perform well with multidimensional data such as images and speech signals, in fact is in the domain of images
that these networks have been used in a physic informed network framework. 

For more on these topic the reader can look at 
\cite{LeC2015_DeepLearning_BenLBH, Che2018_RiseDeepLearning_EngCEW, Muh2021_DeepLearningApplication_AseMAC, Ald2020_DeepLearningApproaches_DerADE, Ber2019_SurveyDeepLearning_BucBBCC, Cal2020_ConvolutionalNetworks_Cal}

\paragraph{CNN architectures}  

Because CNNs were originally created for image recognition, they are better suited to handling image-like data and may not be directly applicable to scientific computing problems, as most geometries are irregular with non-uniform grids; for example, Euclidean-distance-based convolution filters lose their invariance on non-uniform meshes.

A physics-informed geometry-adaptive convolutional neural network (PhyGeoNet) was introduced in \cite{Gao2021_PhygeonetPhysicsInformed_SunGSW}.
PhyGeoNet is a  physics-informed CNN that uses coordinate transformation to convert solution fields from an irregular physical domain to a rectangular reference domain.
Additionally, boundary conditions are strictly enforced making it a physics-constrained neural network.
\\
\noindent
\cite{Fan2021_HighEfficientHybrid_Fan}  observes that a Laplacian operator has been discretized in a convolution operation employing a finite-difference stencil kernel.
Indeed, a Laplacian operator can be discretized using the finite volume approach, and the discretization procedure is equivalent to convolution.
As a result, he enhances PINN by using a finite volume numerical approach with a CNN structure. He devised and implemented a CNN-inspired technique in which, rather than using a Cartesian grid, he computes convolution on a mesh or graph.
Furthermore, rather than zero padding, the padding data serve as the boundary condition.
Finally, \cite{Fan2021_HighEfficientHybrid_Fan}  does not use pooling because the data does not require compression.

\paragraph{Convolutional encoder-decoder network}
%
Autoencoders (AE) (or encoder–decoders) are commonly used to reduce dimensionality in a nonlinear way. 
It consists of two NN components: an encoder that translates the data from the input layer to a finite number of hidden units, and a decoder that has an output layer with the same number of nodes as the input layer
\citep{Che2018_RiseDeepLearning_EngCEW}. 

For modeling stochastic fluid flows, \cite{Zhu2019_PhysicsConstrainedDeep_ZabZZKP}
developed a physics-constrained convolutional encoder-decoder network and a generative model. 
%
\cite{Zhu2019_PhysicsConstrainedDeep_ZabZZKP}  propose a CNN-based technique for solving stochastic PDEs with high-dimensional spatially changing coefficients, demonstrating that it outperforms FC-NN methods in terms of processing efficiency.
\\
\noindent
AE architecture of the convolutional neural network (CNN) is also used in
\cite{Wan2021_TheoryGuidedAuto_ChaWCZ}. 
The authors propose a framework called a Theory-guided Auto-Encoder (TgAE) capable of incorporating physical constraints into the convolutional neural network.
\\
\noindent
\cite{Gen2020_ModelingDynamicsPde_ZabGZ} propose a deep auto-regressive dense encoder-decoder and the physics-constrained training algorithm for predicting transient PDEs. They extend this model to a Bayesian framework to quantify both epistemic and aleatoric uncertainty. 
Finally, \cite{Gru2021_DeepNeuralNetwork_HajGHL} also used an encoder–decoder fully convolutional neural network.

\subsubsection{Recurrent neural networks}
Recurrent neural network (RNN) is a further ANN type, where unlike feed-forward NNs, neurons in the same hidden layer are connected to form a directed cycle. RNNs may accept sequential data as input, making them ideal for time-dependent tasks \citep{Che2018_RiseDeepLearning_EngCEW}.
The RNN processes inputs one at a time, using the hidden unit output as extra input for the next element \citep{Ber2019_SurveyDeepLearning_BucBBCC}.
An RNN's hidden units can keep a state vector that holds a memory of previous occurrences in the sequence.

It is feasible to think of RNN in two ways: first, as a state system with the property that each state, except the first, yields an outcome; secondly, as a sequence of vanilla feedforward neural networks, each of which feeds information from one hidden layer to the next. 
For RNNs, according to \cite{Cat2018_SpecificNetworkDescriptions_ChaCC}, the layerwise function $f$ can be written as
$$
f_i (h_{i-1}) = \alpha (W \cdot h_{i-1} + U \cdot x_i + b).
$$
where $\alpha$ is an elementwise nonlinearity (a typical choice for RNN is the tanh function), 
and where  the hidden vector state $h$  evolves according to
a hidden-to-hidden weight matrix $W$, which starts from an input-to-hidden weight matrix
$U$ and a bias vector $b$.

RNNs have also been enhanced with several memory unit types, such as long short time memory (LSTM) and gated recurrent unit (GRU) \citep{Ald2020_DeepLearningApproaches_DerADE}.
Long short-term memory (LSTM) units  have been created to allow RNNs to handle challenges requiring long-term memories, since LSTM units have a structure called a memory cell that stores information. 
\\
\noindent
Each LSTM layer has a set of interacting units, or cells, similar to those found in a neural network. An LSTM is made up of four interacting units: an internal cell, an input gate, a forget gate, and an output gate. The cell state, controlled by the gates, can selectively propagate relevant information throughout the temporal sequence to capture the long short-term time dependency in a dynamical system \citep{Zha2020_PhysicsInformedMulti_LiuZLS}.
\\
\noindent
The gated recurrent unit (GRU) is another RNN unit developed for extended memory;
GRUs are comparable to LSTM, however they contain fewer parameters and are hence easier to train \citep{Ber2019_SurveyDeepLearning_BucBBCC}.


\paragraph{RNN architectures}
\cite{Via2021_EstimatingModelInadequacy_NasVNDY} introduce, in the form of a neural network, a model discrepancy term into a given ordinary differential equation.
Recurrent neural networks are seen as ideal for dynamical systems because they expand classic feedforward networks to incorporate time-dependent responses. 
Because a recurrent neural network applies transformations to given states in a sequence on a periodic basis, it is possible to design a recurrent neural network cell that does Euler integration; in fact, physics-informed recurrent neural networks can be used to perform numerical integration. 
\\
\noindent
\cite{Via2021_EstimatingModelInadequacy_NasVNDY} 
build recurrent neural network cells in such a way that specific numerical integration methods (e.g., Euler, Riemann, Runge-Kutta, etc.) are employed. The recurrent neural network is then represented as a directed graph, with nodes representing individual kernels of the physics-informed model.
The graph created for the physics-informed model can be used to add data-driven nodes (such as multilayer perceptrons) to adjust the outputs of certain nodes in the graph, minimizing model discrepancy.

\paragraph{LSTM architectures}
Physicists have typically employed distinct LSTM networks to depict the sequence-to-sequence input-output relationship; however, these networks are not homogeneous and cannot be directly connected to one another.
In \cite{Zha2020_PhysicsInformedMulti_LiuZLS} this relationship is expressed using a single network and a central finite difference filter-based numerical differentiator.
\cite{Zha2020_PhysicsInformedMulti_LiuZLS} show two architectures 
for representation learning of sequence-to-sequence features from limited data that is augmented by physics models.
The proposed networks 
is made up of two ($PhyLSTM^2$) or three ($PhyLSTM^3$) deep LSTM networks that describe state space variables, nonlinear restoring force, and hysteretic parameter. Finally, a tensor differentiator, which determines the derivative of state space variables, connects the LSTM networks.
\\
\noindent
Another approach is \cite{Yuc2021_HybridPhysicsInformed_ViaYV}
for temporal integration, that implement an LSTM using a previously introduced Euler integration cell.

\subsubsection{Other architectures for PINN}

Apart from fully connected feed forward neural networks, convolutional neural networks, and recurrent neural networks, this section discusses other approaches that have been investigated. While there have been numerous other networks proposed in the literature, we discovered that only Bayesian neural networks (BNNs) and generative adversarial networks (GANs) have been applied to PINNs. Finally, an interesting application is to combine multiple PINNs, each with its own neural network.


\paragraph{Bayesian Neural Network}
In the Bayesian framework, \cite{Yan2021_BPinnsBayesian_MenYMK} propose to use Bayesian neural networks (BNNs), in their B-PINNs, that consists of a Bayesian neural network subject to the PDE constraint that acts as a prior. 
BNN are neural networks with weights that are distributions rather than deterministic values, and these distributions are learned using Bayesian inference.  
For estimating the posterior distributions, the B-PINN authors use  the Hamiltonian Monte Carlo (HMC) method and the variational inference (VI). 
\cite{Yan2021_BPinnsBayesian_MenYMK} find that for the posterior estimate of B-PINNs, HMC is more appropriate than VI with mean field Gaussian approximation. 
\\
\noindent
They analyse also the possibility to use the  Karhunen-Loève expansion as a stochastic process representation, instead of BNN.
Although the KL is as accurate as BNN and considerably quicker, it cannot be easily applied to high-dimensional situations. 
Finally, they observe that to estimate the posterior of a Bayesian framework, KL-HMC or deep normalizing flow (DNF) models can be employed.
While DNF is more computationally expensive than HMC, it is more capable of extracting independent samples from the target distribution after training. 
This might be useful for data-driven PDE solutions, however it is only applicable to low-dimensional problems.

\paragraph{GAN architectures}
In generative adversarial networks (GANs), two neural networks compete in a zero-sum game to deceive each other.
One network generates and the other discriminates.
The generator accepts input data and outputs data with realistic characteristics.
The discriminator compares the real input data to the output of the generator.
After training, the generator can generate new data that is indistinguishable from real data
\citep{Ber2019_SurveyDeepLearning_BucBBCC}.

\cite{Yan2020_PhysicsInformedGenerative_ZhaYZK} propose a new class of generative adversarial networks (PI-GANs) to address forward, inverse, and mixed stochastic problems in a unified manner. Unlike typical GANs, which rely purely on data for training, PI-GANs use automatic differentiation to embed the governing physical laws in the form of stochastic differential equations (SDEs) into the architecture of PINNs. 
The discriminator in PI-GAN is represented by a basic FFNN, while the generators are a combination of FFNNs and a NN induced by the SDE. 

\paragraph{Multiple PINNs} 
A final possibility is to combine several PINNs, each of which could be implemented using a different neural network. 
\cite{Jag2020_ConservativePhysicsInformed_KhaJKK}
propose a conservative physics-informed neural network (cPINN) on discrete domains. In this framework, the complete solution is recreated by patching together all of the solutions in each sub-domain using the appropriate interface conditions. 
This type of domain segmentation also allows for easy network parallelization, which is critical for obtaining computing efficiency.
This method may be expanded to a more general situation, called by the authors as Mortar PINN, for connecting non-overlapping deconstructed domains.
Moreover, the suggested technique may use totally distinct neural networks, in each subdomain, with various architectures to solve the same underlying PDE.
\\
\noindent
\cite{Sti2020_LargeScaleNeural_BetSBB} proposes the GatedPINN architecture by incorporating conditional computation into PINN.
This architecture design is composed of a  gating network and set of PINNs, hereinafter referred to as ``experts''; each expert solves the problem for each space-time point, and their results are integrated via a gating network.
The gating network determines which expert should be used and how to combine them.
We will use one of the expert networks as an example in the following section of this review.

\begin{table}[hbt!]
\centering
\resizebox{\textwidth}{!}{%
\begin{tabular}{@{} c  l  l  @{}}
\toprule
\textbf{NN family} &
  \textbf{NN type} &
  \textbf{Papers} \\ \midrule
\multirow{6}{*}{FF-NN} &
  1 layer / EML &
  \begin{tabular}[c]{@{}l@{}}\cite{Dwi2020_PhysicsInformedExtreme_SriDS},\\ \cite{Sch2021_ExtremeTheoryFunctional_FurSFL}\end{tabular} \\ \cline{2-3} 
 &
  2-4 layers &
  \begin{tabular}[c]{@{}l@{}}32 neurons per layer \cite{He2020_PhysicsInformedNeural_BarHBTT} \\ 50 neurons per layer \cite{Tar2020_PhysicsInformedDeep_MarTMP}\end{tabular} \\ \cline{2-3} 
 &
  5-8 layers &
  250 neurons per layer \cite{Zhu2021_MachineLearningMetal_LiuZLY} \\ \cline{2-3} 
 &
  9+ layers &
  \begin{tabular}[c]{@{}l@{}}\cite{Che2021_DeepLearningMethod_ZhaCZ}\\ \cite{Wah2021_PinneikEikonalSolution_HagWHA}\end{tabular} \\ \cline{2-3} 
 &
  Sparse &
  \cite{Ram2021_SpinnSparsePhysics_RamRR} \\ \cline{2-3} 
 &
  multi FC-DNN &
  \begin{tabular}[c]{@{}l@{}}\cite{Ami2021_PhysicsInformedNeural_HagANHC} \\ \cite{Isl2021_ExtractionMaterialProperties_ThaITMH} \end{tabular} \\ \midrule
\multirow{2}{*}{CNN} &
  plain CNN &
  \begin{tabular}[c]{@{}l@{}}\cite{Gao2021_PhygeonetPhysicsInformed_SunGSW}\\ \cite{Fan2021_HighEfficientHybrid_Fan} 
  \end{tabular} \\ \cline{2-3} 
 &
  AE CNN &
  \begin{tabular}[c]{@{}l@{}}\cite{Zhu2019_PhysicsConstrainedDeep_ZabZZKP}, \\ \cite{Gen2020_ModelingDynamicsPde_ZabGZ} \\ \cite{Wan2021_TheoryGuidedAuto_ChaWCZ}\end{tabular} \\ \midrule
\multirow{2}{*}{RNN} &
  RNN &
  \cite{Via2021_EstimatingModelInadequacy_NasVNDY} \\ \cline{2-3} 
 &
  LSTM &
  \begin{tabular}[c]{@{}l@{}}\cite{Zha2020_PhysicsInformedMulti_LiuZLS}\\ \cite{Yuc2021_HybridPhysicsInformed_ViaYV}\end{tabular} \\ \midrule
\multirow{2}{*}{Other} &
  BNN &
  \cite{Yan2021_BPinnsBayesian_MenYMK} \\ \cline{2-3} 
 &
  GAN &
  \cite{Yan2020_PhysicsInformedGenerative_ZhaYZK} \\ \bottomrule
\end{tabular}%
}
\caption{The main neural network utilized in PINN implementations is synthesized in this table.
We summarize Section 2 by showcasing some of the papers that represent each of the many Neural Network implementations of PINN.
Feedforward neural networks (FFNNs), convolutional neural networks (CNNs), and recurrent neural networks (RNN) are the three major families of Neural Networks reported here.
A publication is reported for each type that either used this type of network first or best describes its implementation.
In the literature, PINNs have mostly been implemented with FFNNs with 5-10 layers.
CNN appears to have been applied in a PCNN manner for the first time, by incorporating the boundary condition into the neural network structure rather than the loss. 
}
\label{tab:NeuralNetworks}
\end{table}

\subsection{Injection of Physical laws}

To solve a PDE with PINNs, derivatives of the network's output with respect to the inputs are needed.
Since the function $\bm{u}$ is approximated by a NN with smooth activation function,
$\hat{\bm{u}}_\theta$, it
can be differentiated. 
There are four methods for calculating derivatives: hand-coded, symbolic, numerical, and automatic.
Manually calculating derivatives may be correct, but it is not automated and thus impractical \citep{Wah2021_PinneikEikonalSolution_HagWHA}.

Symbolic and numerical methods like finite differentiation perform very badly when applied to complex functions; automatic differentiation (AD), on the other hand, overcomes numerous restrictions as floating-point precision errors, for numerical differentiation, or memory intensive symbolic approach.
AD use exact expressions with floating-point values rather than symbolic strings, there is no approximation error \citep{Wah2021_PinneikEikonalSolution_HagWHA}.

Automatic differentiation (AD) is also known as autodiff, or algorithmic differentiation, although it would be better to call it algorithmic differentiation since AD does not totally automate differentiation: instead of symbolically evaluating derivatives operations, AD performs an analytical evaluation of them.

Considering  a function
$f:\mathbb{R}^n\to\mathbb{R}^m$ of which we want to calculate the Jacobian $J$,  after calculating the graph of all the operations composing the mathematical expression, AD can then work in either forward or reverse mode for calculating the numerical derivative.



AD results being the main  technique used in literature and used by all PINN implementations, in particular only 
\cite{Fan2021_HighEfficientHybrid_Fan}
use local fitting method
approximation of the differential operator to solve the PDEs instead of automatic differentiation (AD).
Moreover, by using local fitting method rather than employing automated differentiation, Fang is able to verify that a his PINN implementation has a convergence.


Essentially, AD incorporates a PDE into the neural network's loss equation~\eqref{eq:general_loss}, where
the differential equation residual is 
$$
r_\mathcal{F}[ \hat{\bm{u}}_{\theta}](\bm{z}) = r_{\theta}(\bm{z}):= \mathcal{F}( \hat{\bm{u}}_{\theta} (\bm{z}); \gamma  ) -  \bm{f},
$$
and similarly the residual NN corresponding to boundary conditions (BC) or initial conditions (IC) is obtained by substituting $\hat{\bm{u}}_{\theta}$ in the second equation of~\eqref{eq:general_form}, i.e.
$$
r_\mathcal{B}[ \hat{\bm{u}}_{\theta}](\bm{z}) :=
\mathcal{B}(\hat{\bm{u}}_{\theta}(\bm{z})) -  \bm{g}(\bm{z}).
$$
Using these residuals, it is possible to assess how well an approximation $u_{\theta}$ satisfies~\eqref{eq:general_form}.
It is worth noting that for the exact solution, $u$, the residuals are $r_\mathcal{F}[u]=r_\mathcal{B}[u]=0$ \citep{De2022_ErrorEstimatesPhysics_JagDRJM}.

In 
\cite{Rai2018_HiddenPhysicsModels_KarRK, Rai2019_PhysicsInformedNeural_PerRPK}, the original formulation of the aforementioned differential equation residual, leads to the form of
$$
r_\mathcal{F}[ \hat{u}_{\theta}](\bm{z}) =  r_{\theta}(\bm{x},t)= \frac{\partial}{\partial t} \hat{u}_{\theta}(\bm{x},t)  + \mathcal{F}_{\bm{x}}\hat{u}_{\theta}(\bm{x},t).
$$

In the deep learning framework, the principle of imposing physical constraints is represented by differentiating neural networks with respect to input spatiotemporal coordinates using the chain rule. 
In \cite{Mat2021_UncoveringTurbulentPlasma_FraMFH} the model loss functions are embedded and then further normalized into dimensionless form.
The repeated differentiation, with AD, and composition of networks used to create each individual term in the partial differential equations results in a much larger resultant computational graph.
As a result, the cumulative computation graph is effectively an approximation of the PDE equations \citep{Mat2021_UncoveringTurbulentPlasma_FraMFH}.

The chain rule is used in automatic differentiation for several layers to compute derivatives hierarchically from the output layer to the input layer.
Nonetheless, there are some situations in which the basic chain rule does not apply.
\cite{Pan2019_FpinnsFractionalPhysics_LuPLK}  substitute fractional differential operators with their discrete versions, which are subsequently incorporated into the PINNs' loss function. 

\subsection{Model estimation by learning approaches}\label{sec:ModelEstimation}

The PINN methodology determines the parameters $\theta$ of the NN, $\hat{\bm{u}}_\theta$, by minimizing a loss function, i.e.
\begin{equation*}
\theta =  \mathop{\argmin}_{\theta}  \mathcal{L}(\theta) 
\end{equation*}
where
\begin{equation}\label{eq:loss_pinn}
\mathcal{L}(\theta) = 
\omega_\mathcal{F} \mathcal{L}_\mathcal{F}(\theta) +
\omega_\mathcal{B} \mathcal{L}_\mathcal{B}(\theta) +
\omega_{d} \mathcal{L}_{data} (\theta) 
.
\end{equation}
The three terms of $\mathcal{L}$ refer to the errors in describing the initial $\mathcal{L}_{i}$ or boundary condition $\mathcal{L}_{b}$, both indicated as $\mathcal{L}_\mathcal{B}$, the loss respect the partial differential equation $\mathcal{L}_{\mathcal{F}}$, and the validation of known data points $\mathcal{L}_{data}$. 
Losses are usually defined in the literature as a sum, similar to the previous equations, however they can be considered as integrals
\begin{equation*}
\mathcal{L}_\mathcal{F}(\theta) 
= 
\int_{\bar{\Omega}} \left( \mathcal{F} (\hat{\bm{u}}_\theta(\bm{z}))   - \bm{f}(\bm{z}_i)) \right)^2 \, d\bm{z}
\end{equation*}
This formulation is not only useful for a theoretical study, as we will see in~\ref{sec:Theory}, but it is also implemented in a PINN package, NVIDIA Modulus \citep{Modulus2021}, allowing for more effective integration strategies, such as sampling with higher frequency in specific areas of the domain to more efficiently approximate the integral losses.

Note that, if the PINN framework is employed as a supervised methodology, the neural network parameters are chosen by minimizing the difference between the observed outputs and the model's predictions; otherwise, just the PDE residuals are taken into account. 

As in equation~\eqref{eq:loss_pinn} the first term, $\mathcal{L}_\mathcal{F}$, represents the loss produced by a mismatch with the governing differential equations $\mathcal{F}$ \citep{He2020_PhysicsInformedNeural_BarHBTT, Sti2020_LargeScaleNeural_BetSBB}.
It enforces the differential equation $\mathcal{F}$ at the \emph{collocation points}, which can be chosen uniformly or unevenly over the domain $\Omega$ of equation~\eqref{eq:general_form}.
\\
\noindent
The remaining two losses attempt to fit the known data over the NN.
The loss caused by a mismatch with the data (i.e., the measurements of $\bm{u}$) is denoted by $\mathcal{L}_{data} (\theta)$.
The second term typically forces $\hat{\bm{u}}_\theta$ to mach the measurements of $\bm{u}$ over provided points $(\bm{z}, \bm{u}^*)$, which can be given as synthetic data or actual measurements, and the weight $\omega_d$ can account for the quality of such measurements.
\\
\noindent
The other term is the loss due to mismatch with the boundary or initial conditions, $\mathcal{B} (\hat{\bm{u}}_\theta) = \bm{g}$ from equation~\eqref{eq:general_form}.
%

Essentially, the training approach recovers the shared network parameters $\theta$ from:
\begin{itemize}
    \item few scattered observations of $\bm{u}(\bm{z})$, specifically $\{\bm{z}_{i}, \bm{u}_i^*\}$, $i = 1,\dots,N_d$
    \item as well as a greater number of collocation points $\{\bm{z}_{i}, \bm{r}_i = 0\}$, $i = 1,\dots,N_r$ for the residual,
\end{itemize}

The resulting optimization problem can be handled using normal stochastic gradient descent without the need for constrained optimization approaches by minimizing the combined loss function.
A typical implementation of the loss uses a mean square error formulation \citep{Kol2021_PhysicsInformedNeural_DAKDJH}, where:
\begin{equation*}
\mathcal{L}_\mathcal{F}(\theta) =  MSE_\mathcal{F} 
= \frac{1}{N_c} \sum_{i=1}^{N_c} \|
\mathcal{F} (\hat{\bm{u}}_\theta(\bm{z}_i))   - \bm{f}(\bm{z}_i))
\|^2    
= \frac{1}{N_c} \sum_{i=1}^{N_c} \|r_{\theta}(\bm{u}_{i}) - \bm{r}_i\|^2
\end{equation*}
enforces the PDE on a wide set of randomly selected collocation locations inside the domain, i.e. penalizes the difference between the estimated left-hand side of a PDE and the known right-hand side of a PDE  \citep{Kol2021_PhysicsInformedNeural_DAKDJH}; %
other approaches may employ an integral definition of the loss \citep{Hen2021_NvidiaSimnetAi_NarHNN}.
As for the boundary and initial conditions, instead 
\begin{equation*}
\mathcal{L}_\mathcal{B}(\theta)  = MSE_\mathcal{B} 
= \frac{1}{N_B} \sum_{i=1}^{N_B} 
\|
\mathcal{B} (\hat{\bm{u}}_\theta(\bm{z}))   - \bm{g}(\bm{z}_i))
\|^2    
\end{equation*}
whereas for the data points,
\begin{equation*}
\mathcal{L}_{data}(\theta)  = MSE_{data}  = \frac{1}{N_d}\sum\limits_{i=1}^{N_d}\|\hat{\bm{u}}_{\theta}(\bm{z}_{i}) - \bm{u}_i^*\|^2 .
\end{equation*}
computes the error of the approximation $u(t, x)$ at known data points. In the case of a forward problem, the data loss might also indicate the boundary and initial conditions, while in an inverse problem it refers to the solution at various places inside the domain  \citep{Kol2021_PhysicsInformedNeural_DAKDJH}.


In 
\cite{Rai2018_HiddenPhysicsModels_KarRK, Rai2019_PhysicsInformedNeural_PerRPK}, original approach the overall loss~\eqref{eq:loss_pinn} was formulated as
$$
\mathcal{L}(\theta) =   \frac{1}{N_c} \sum_{i=1}^{N_c} \| \frac{\partial}{\partial t} \hat{u}_{\theta}(\bm{x},t)  + \mathcal{F}_{\bm{x}}\hat{u}_{\theta}(\bm{x},t) - \bm{r}_i\|^2
 +
 \frac{1}{N_d}\sum\limits_{i=1}^{N_d}\|\hat{\bm{u}}_{\theta}(x_{i},t_i) - \bm{u}_i^*\|^2 .
$$
 
The gradients in $ \mathcal{F}$ are derived using automated differentiation.
The resulting predictions are thus driven to inherit any physical attributes imposed by the PDE constraint \citep{Yan2019_AdversarialUncertaintyQuantification_PerYP}.  
The physics constraints are included in the loss function to enforce model training, which can accurately reflect latent system nonlinearity even when training data points are scarce \citep{Zha2020_PhysicsInformedMulti_LiuZLS}.

\paragraph{Observations about the loss}

The loss $\mathcal{L}_\mathcal{F}(\theta)$ is calculated by utilizing automated differentiation (AD) to compute the derivatives of $\hat{\bm{u}}_\theta(\bm{z})$ \citep{He2020_PhysicsInformedNeural_BarHBTT}.
Most ML libraries, including TensorFlow and Pytorch, provide AD, which is mostly used to compute derivatives with respect to DNN weights (i.e. $\theta$).
AD permits the PINN approach to implement any PDE and boundary condition requirements without numerically discretizing and solving the PDE \citep{He2020_PhysicsInformedNeural_BarHBTT}.

Additionally, by applying PDE constraints via the penalty term $\mathcal{L}_\mathcal{F}(\theta)$, it is possible to use the related weight $\omega_\mathcal{F}$ to account for the PDE model's fidelity. 
To a low-fidelity PDE model, for example, can be given a lower weight.
In general, the number of unknown parameters in $\theta$ is substantially greater than the number of measurements, therefore regularization is required for DNN training \citep{He2020_PhysicsInformedNeural_BarHBTT}.

By removing loss for equations from the optimization process (i.e., setting $\omega_\mathcal{F}=0$), neural networks could be trained without any knowledge of the underlying governing equations.
Alternatively, supplying initial and boundary conditions for all dynamical variables would correspond to solving the equations directly with neural networks on a regular basis \citep{Mat2021_UncoveringTurbulentPlasma_FraMFH}.

While it is preferable to enforce the physics model across the entire domain, the computational cost of estimating and reducing the loss function~\eqref{eq:loss_pinn}, while training,  grows with the number of residual points \citep{He2020_PhysicsInformedNeural_BarHBTT}.  
Apart the number of residual points, also the position (distribution) of residual points are crucial parameters in PINNs because they can change the design of the loss function \citep{Mao2020_PhysicsInformedNeural_JagMJK}.

A deep neural network can reduce approximation error by increasing network expressivity, but it can also produce a large generalization error.
Other hyperparameters, such as learning rate, number of iterations, and so on, can be adjusted to further control and improve this issue. 


The addition of extra parameters layer by layer  in a NN modifies the slope of the activation function in each hidden-layer, improving the training speed.
Through the slope recovery term, these activation slopes can also contribute to the loss function \citep{Jag2020_ConservativePhysicsInformed_KhaJKK}.

\paragraph{Soft and hard constraint}

BC constraints can be regarded as penalty terms (soft BC enforcement) \citep{Zhu2019_PhysicsConstrainedDeep_ZabZZKP},
 or they can be encoded into the network design (hard BC enforcement) \citep{Sun2020_SurrogateModelingFluid_GaoSGPW}.
%
Many existing PINN frameworks use a \emph{soft} approach to constrain the BCs by creating extra loss components defined on the collocation points of borders.
The disadvantages of this technique are twofold:
\begin{enumerate}
    \item  satisfying the BCs accurately is not guaranteed;
    \item the assigned weight of BC loss might effect learning efficiency, and no theory exists to guide determining the weights at this time.
\end{enumerate}
%
\cite{Zhu2021_MachineLearningMetal_LiuZLY} address the Dirichlet BC in a \emph{hard} approach by employing a specific component of the neural network to purely meet the specified Dirichlet BC. 
Therefore, the initial boundary conditions are regarded as part of the labeled data constraint.


When compared to the residual-based loss functions typically found in the literature, the variational energy-based loss function is simpler to minimize and so performs better
\citep{Gos2020_TransferLearningEnhanced_AniGACR}.
%
Loss function can be constructed using
collocation points,
weighted residuals derived by the Galerkin-Method \citep{Kha2019_VariationalPhysicsInformed_ZhaKZK},
or energy based.
%
Alternative loss functions approaches are compared in \cite{Li2021_PhysicsGuidedNeural_BazLBZ}, by using either only data-driven (with no physics model), a PDE-based loss, and an energy-based loss. 
They observe that there are advantages and disadvantages for both PDE-based and  energy-based approaches. PDE-based loss function has more hyperparameters than the energy-based loss function. The energy-based strategy is more sensitive to the size and resolution of the training samples than the PDE-based strategy, but it is more computationally efficient.

\paragraph{Optimization methods}

The minimization process of the loss function is called \emph{training};
in most of the PINN literature, loss functions are optimized using minibatch sampling using 
Adam and the limited-memory Broyden-Fletcher-Goldfarb-Shanno (L-BFGS) algorithm, a quasi-Newton optimization algorithm. When monitoring noisy data, \cite{Mat2021_UncoveringTurbulentPlasma_FraMFH} found that increasing the sample size and training only with L-BFGS achieved the optimum for learning.

For a moderately sized NN, such as one with four hidden layers (depth of the NN is 5) and twenty neurons in each layer (width of the NN is 20), we have over 1000 parameters to optimize.
There are several local minima for the loss function, and the gradient-based optimizer will almost certainly become caught in one of them; finding global minima is an NP-hard problem \citep{Pan2019_FpinnsFractionalPhysics_LuPLK}.

The Adam approach, which combines adaptive learning rate and momentum methods, is employed in \cite{Zhu2021_MachineLearningMetal_LiuZLY} to increase convergence speed, because stochastic gradient descent (SGD) hardly manages random collocation points, especially in 3D setup.

\cite{Yan2020_PhysicsInformedGenerative_ZhaYZK} use Wasserstein GANs with gradient penalty (WGAN-GP) and prove that they are more stable than vanilla GANs, in particular for approximating stochastic processes with deterministic boundary conditions.

cPINN \citep{Jag2020_ConservativePhysicsInformed_KhaJKK} allow to flexibly select network hyper-parameters such as optimization technique, activation function, network depth, or network width based on intuitive knowledge of solution regularity in each sub-domain.
E.g. for smooth zones, a shallow network may be used, and a deep neural network can be used in an area where a complex nature is assumed.

\cite{He2020_PhysicsInformedNeural_BarHBTT}  propose a two-step training approach in which the loss function is minimized first by the Adam algorithm with a predefined stop condition, then by the L-BFGS-B optimizer.
According to the aforementioned paper, for cases with a little amount of training data and/or residual points, L-BFGS-B, performs better with a faster rate of convergence and reduced computing cost. 




Finally, let's look at a practical examples for optimizing the training process; dimensionless and normalized data are used in DeepONet training to improve stability \citep{Lin2021_SeamlessMultiscaleOperator_MaxLMLK}. 
Moreover the governing equations in dimensionless form, including the Stokes equations, electric potential, and ion transport equations, are presented in DeepM\&Mnets  \citep{Cai2021_DeepmmnetInferringElectroconvection_WanCWL}.
%

In terms of training procedure initialization, slow and fast convergence behaviors are produced by bad and good initialization, respectively, but \cite{Pan2019_FpinnsFractionalPhysics_LuPLK}  reports a technique for selecting the most suitable one.
By using a limited number of iterations, one can first solve the inverse problem.
This preliminary solution has low accuracy due to discretization, sampling, and optimization errors. The optimized parameters from the low-fidelity problem can then be used as a suitable initialization.

\subsection{Learning theory of PINN }\label{sec:Theory}

This final subsection provides most recent theoretical studies on PINN to better understand how they work and their potential limits.
These investigations are still in their early stages, and much work remains to be done.

Let us start by looking at how PINN can approximate the true solution of a differential equation, similar to how error analysis is done a computational framework. 
In traditional numerical analysis, we approximate the true solution $\bm{u}(\bm{z})$  of a problem with an approximation scheme that computes $\hat{\bm{u}}_\theta(\bm{z})$. The main theoretical issue is to estimate the global error 
$$\mathcal{E} = \hat{\bm{u}}_\theta(\bm{z}) - \bm{u}(\bm{z}).$$ 
Ideally, we want to find a set of parameters, $\theta$, such that $\mathcal{E}=0$.

When solving differential equations using a numerical discretization technique, we are interested in the numerical method's stability, consistency, and convergence \citep{Rya2006_TheoreticalIntroductionNumerical_TsyRT, Arn2015_StabilityConsistencyConvergence_Arn, Tho1992_NumericalMethods101convergence_Tho}.
%
%
\\
In such setting, discretization's error can be bound in terms of consistency and stability, a basic result in numerical analysis.
The Lax-Richtmyer Equivalence Theorem is often referred to as a fundamental result of numerical analysis 
where 
roughly the convergence is ensured when there is consistency and stability.

%
When studying PINN to mimic this paradigm, the convergence and stability are related to how well the NN learns from physical laws and data.
In this conceptual framework, we use a NN, which is a parameterized approximation of problem solutions modeled by physical laws.
In this context, we will (i) introduce the concept of convergence for PINNs, (ii) revisit the main error analysis definitions in a statistical learning framework, and (iii) finally report results for the generalization error. 

%

\subsubsection{Convergence aspects}

The goal of a mathematical foundation for the PINN theory is to investigate the convergence of the computed $\hat{\bm{u}}_\theta (\bm{z})$ to the solution of problem~\eqref{eq:general_form}, $\bm{u}(\bm{z})$.

Consider a NN configuration with coefficients compounded in the vector  $\bm{\theta}$ and a cardinality equal to the number of coefficients of the NN, $\#\bm{\theta}$.
In such setting, we can consider the hypothesis class
$$
\mathcal{H}_n = \{ \bm{u}_\theta  : \#\bm{\theta} = n \}
$$
composed of all the predictors representing a  NN  whose number of coefficients of  the architecture is $n$.
The capacity of a PINN to be able to learn, is related to how big is $n$, i.e. the expressivity  of $\mathcal{H}_n$.

In such setting, a theoretical issue, is to investigate, how dose the sequence of compute predictors, $\hat{u}_\theta$, converges to the solution of the physical problem ~\eqref{eq:general_form}
$$
\hat{u}_{\theta(n)} \to   u  ,\qquad n\to \infty.
$$

A recent result in this direction was  obtained  by \cite{De2021_ApproximationFunctionsTanh_LanDRLM}
in which they proved that the difference $\hat{u}_{\theta(n)} - u$ converges to zero as the width of a predefined NN, with activation function tanh,  goes to infinity.
\\
Practically, the PINN requires choosing a network class $\mathcal{H}_n$  and a loss function given a collection of $N$-training data \citep{Shi2020_ConvergencePhysicsInformed_DarSDK}.
Since the quantity and quality of training data affect  $\mathcal{H}_n$, 
the goal is to minimize the loss, by finding a $u_{\theta^*} \in \mathcal{H}_n$, by training the $N$ using an optimization process.
Even if $\mathcal{H}_n$ includes the exact solution $u$ to PDEs and a global minimizer is established, there is no guarantee that the minimizer and the solution $u$ will coincide.
A first work related on PINN \citep{Shi2020_ConvergencePhysicsInformed_DarSDK}, the authors show that the sequence of minimizers $\hat{\bm{u}}_{\theta^*}$ strongly converges to the solution of a linear second-order elliptic and parabolic type PDE.

\subsubsection{Statistical Learning error analysis}

The entire learning process of a PINN can be considered as a statistical learning problem, and it involves mathematical foundations aspects for the error analysis \citep{Kut2022_MathematicsArtificialIntelligence_Kut}.
%
%
For a mathematical treatment of errors in PINN, it is important to take into account: optimization, generalization errors, and approximation error.
It is worth noting that the last one is dependent on the architectural design.  
\\
Let be $N$ collocation points on $\bar{\Omega} = \Omega \cup \partial \Omega$,
a  NN approximation realized with $\theta$, denoted by $\hat{u}_\theta$, evaluated at points $z_i$, whose exact value is $h_i$.
Following the notation of \cite{Kut2022_MathematicsArtificialIntelligence_Kut}, the \emph{empirical risk} is defined as
\begin{equation}\label{eq:risk1} 
\widehat{\mathcal{R}}[u_\theta] := \frac{1}{N} \sum_{i=1}^{N} \|
\hat{u}_\theta(z_i) - h_i,
\|^2    
\end{equation}
and represents how well the NN is able to predict the exact value of the problem. The  empirical risk actually corresponds to the loss defined in~\ref{sec:ModelEstimation}, where $\hat{\bm{u}}_\theta = \mathcal{F} (\hat{\bm{u}}_\theta(\bm{z}_i))$  and $h_i=\bm{f}(\bm{z}_i)$ and similarly for the boundaries.

A continuum perspective is the \emph{risk} of using an approximator  $\hat{u}_\theta$, calculated as follows: 
\begin{equation} \label{eq:risk}
\mathcal{R}[\hat{u}_\theta] := \int_{\bar{\Omega}} (\hat{u}_\theta(z) - u(z))^2 \, dz,
\end{equation}
where the distance between the approximation $\hat{u}_\theta$ and the solution $u$ is obtained with the $L^2$-norm.
The final approximation computed by the PINN, after a training process of DNN, is $\hat{u}_\theta^*$. 
The main aim in error analysis, is to find suitable estimate for the risk of predicting $u$ i.e. $\mathcal{R}[\hat{u}_\theta^*]$.
\\
The training process, uses gradient-based optimization techniques to minimize a generally non convex cost function. 
In practice, the algorithmic optimization scheme will not always find a global minimum. 
So the error analysis takes into account the \emph{optimization error} defined as follows:
\begin{equation*}
\mathcal{E}_O :=
\widehat{\mathcal{R}}[\hat{u}_\theta^*] - \inf_{\theta \in \Theta} \widehat{\mathcal{R}}[u_\theta]
\end{equation*}
\\
Because the objective function is nonconvex, the optimization error is unknown.
Optimization frequently involves a variety of engineering methods and time-consuming fine-tuning, using, gradient-based optimization methods.
Several stochastic gradient descent methods have been proposed,and
many PINN use Adam and L-BFGS.
Empirical evidence suggests that gradient-based optimization techniques perform well in different challenging tasks; however, gradient-based optimization might not find a global minimum for many ML tasks, such as for PINN, and this is still an open problem \citep{Shi2020_ConvergencePhysicsInformed_DarSDK}.

Moreover a measure of the prediction accuracy on unseen data in machine learning is the 
\emph{generalization error}:
\begin{equation*}
\mathcal{E}_G :=
\sup_{\theta \in \Theta} 
\lvert \mathcal{R}[u_\theta]  -\widehat{\mathcal{R}}[u_\theta] \rvert 
\end{equation*}
\\
The generalization error measures how well the loss integral is approximated in relation to a specific trained neural network. 
One of the first  paper focused with convergence of generalization error is \cite{Shi2020_ConvergencePhysicsInformed_DarSDK}.

About the ability of the NN to approximate the exact solution, the \emph{approximation error} is defined as
\begin{equation*}
\mathcal{E}_A :=
\inf_{\theta \in \Theta}\mathcal{R}[u_\theta]
\end{equation*}
\\
The approximation error is well studied in general, in fact we know that
one layer neural network with a high width can evenly estimate a function and its partial derivative as shown by \cite{Pin1999_ApproximationTheoryMlp_Pin}.

Finally, as stated in \cite{Kut2022_MathematicsArtificialIntelligence_Kut}, 
the global error between the trained deep neural network $\hat{u}_\theta^*$ and the correct solution function $u$ of problem~\eqref{eq:general_form}, can so be bounded by the previously defined error in the following way
\begin{equation} \label{eq:error}
\mathcal{R}[\hat{u}_\theta^*] 
\leq
\mathcal{E}_O
+
2 \mathcal{E}_G
+
\mathcal{E}_A
\end{equation}


These considerations lead to the major research threads addressed in recent studies, which are currently being investigated for PINN and DNNs in general  \cite{Kut2022_MathematicsArtificialIntelligence_Kut}.

\subsubsection{Error analysis results for PINN}


About the approximating error,
since it depends on the NN architecture, mathematical foundations results are generally discussed in papers deeply focused on this topic \cite{Cal2020_UniversalApproximators_Cal, Elb2021_DeepNeuralNetwork_PerEPGB}.

However, a first argumentation strictly related to PINN is reported in \cite{Shi2020_ErrorEstimatesResidual_ZhaSZK}.
One of the main theoretical results 
on $\mathcal{E}_A$, can be found in \cite{De2021_ApproximationFunctionsTanh_LanDRLM}.
They demonstrate that for a neural network with a tanh activation function and only two hidden layers, $\hat{u}_\theta$,  may approximate a function $u$ with a bound in a Sobolev space:
\begin{equation*}
\|\hat{u}_{\theta_N} - u\|_{W^{k,\infty}}  
\leq
C
\frac{\ln (cN)^k}{N^{s-k}}
\end{equation*}
where $N$ is the number of training points, $c,C>0$ are constants independent of $N$ and explicitly known, $u\in W^{ s,\infty} ([0,1]^d)$.
We remark that the NN has width $N^d$, and $\#\theta$ depends on both the number of training points $N$ and the dimension of the problem $d$.
\\




Formal findings for generalization errors in PINN are provided specifically for a certain class of PDE.
In \cite{Shi2020_ConvergencePhysicsInformed_DarSDK} they provide convergence estimate for linear second-order elliptic and parabolic type PDEs, while in 
\cite{Shi2020_ErrorEstimatesResidual_ZhaSZK} they extend the results to all linear problems,  including hyperbolic equations.
\cite{Mis2022_EstimatesGeneralizationError_MolMM} gives an abstract framework for PINN on forward problem for PDEs, they estimate the generalization error by means of training error (empirical risk), and number of training points, such abstract framework is also addressed for inverse problems \citep{Mis2021_EstimatesGeneralizationError_MolMM}.
In \cite{De2022_ErrorEstimatesPhysics_JagDRJM}
the authors specifically address Navier-Stokes equations and show that
small training error imply a small generalization error, by proving that
\begin{equation*}
\mathcal{R}[\hat{u}_\theta] =  \|u - \hat{u}_{\theta}\|_{L^2}  
\leq
\left(
C
\widehat{\mathcal{R}}[u_\theta] 
+ 
\mathcal{O}\left(N^{-\frac{1}{d}}\right)
\right)^{\frac{1}{2}}.
\end{equation*}
This estimate suffer from the curse of dimensionality (CoD), that is to say, in order to reduce the error by a certain factor, the number of training points needed and the size of the neural network, scales up exponentially.



\cite{De2021_ErrorAnalysisPhysics_MisDRM} prove that for a Kolmogorov type PDE (i.e. heat equation or Black-Scholes equation), 
the following inequality holds, almost always,
\begin{equation*}
\mathcal{R}[\hat{u}_\theta]
\leq
\left(
C
\widehat{\mathcal{R}}[u_\theta] 
+ 
\mathcal{O}\left(N^{-\frac{1}{2}}\right)
\right)^{\frac{1}{2}},
\end{equation*}
and is not dependant on the dimension of the problem $d$.

Finally,
\cite{Mis2021_PhysicsInformedNeural_MolMM}
investigates the radiative transfer equation,  which is noteworthy for its high-dimensionality, with the radiative intensity being a function of 7 variables (instead of 3, as common in many physical problems).
The authors prove also here that the generalization error is bounded by the training error and the number of training points, and the dimensional dependence is on a logarithmic factor:
\begin{equation*}
\mathcal{R}[\hat{u}_\theta]
\leq
\left(
C
\widehat{\mathcal{R}}[u_\theta] ^2
+ 
c\left( \frac{(\ln N)^{2d}}{N}    \right)
\right)^{\frac{1}{2}}.
\end{equation*}
The authors are able to show that PINN does not suffer from the dimensionality curse for this problem, observing that the training error does not depend on the dimension but only on the number of training points.

\section{Differential problems dealt with PINNs}\label{sec4}

The first vanilla PINN \citep{Rai2019_PhysicsInformedNeural_PerRPK} was built to solve complex nonlinear PDE equations of the form $\bm{u}_t + \mathcal{F}_{\bm{x}}\bm{u} = 0$,
 where $\bm{x}$  is a vector of space coordinates, $t$ is a vector time coordinate, and $\mathcal{F}_{\bm{x}}$ is a nonlinear differential operator with respect to spatial coordinates.
First and mainly, the PINN architecture was shown to be capable of handling both forward and inverse problems.
Eventually, in the years ahead, PINNs have been employed to solve a wide variety of ordinary differential equations (ODEs), partial differential equations (PDEs), Fractional PDEs, and integro-differential equations (IDEs), as well as stochastic differential equations (SDEs). 
This section is dedicated to illustrate where research has progressed in addressing various sorts of equations, by grouping equations according to their form and addressing the primary work in literature that employed PINN to solve such equation. 
All PINN papers dealing with ODEs will be presented first. 
Then, works on steady-state PDEs such as Elliptic type equations, steady-state diffusion, and the Eikonal equation are reported.
The Navier--Stokes equations are followed by more dynamical problems such as heat transport, advection-diffusion-reaction system, hyperbolic equations, and Euler equations or quantum harmonic oscillator.
Finally, while all of the previous PDEs can be addressed in their respective Bayesian problems, the final section provides insight into how uncertainly is addressed, as in  stochastic equations.


\subsection{Ordinary Differential Equations}
ODEs can be used to simulate complex nonlinear systems  which are difficult to model using simply physics-based models \citep{Lai2021_StructuralIdentificationPhysics_MylLMNC}. A typical ODE system is written as 
\begin{equation*}
\frac{d \bm{u}(\bm{x}, t)}{dt} = f(\bm{u}(\bm{x},t),t)
\end{equation*}
where initial conditions can be specified as $\mathcal{B}(\bm{u}(t))=  \bm{g}(t)$, resulting in an initial value problem or boundary value problem with  $\mathcal{B}(\bm{u}(\bm{x})) = \bm{g}(\bm{x})$. 
A PINN approach is used by  \cite{Lai2021_StructuralIdentificationPhysics_MylLMNC} for structural identification, using Neural Ordinary Differential Equations (Neural ODEs). Neural ODEs can be considered as a continuous representation of ResNets (Residual Networks), by using a neural network to parameterize a  dynamical system in the form of ODE for an initial value problem (IVP): 
\begin{equation*}
\frac{d \bm{u}(t)}{dt} = f_{\theta}(\bm{u}(t),t)  ; \bm{u}(t_0) = \bm{u}_0
\end{equation*}
where $f$ is the neural network parameterized by the vector $\theta$.

The idea is to use Neural ODEs as learners of the governing dynamics of the systems, and so to structure of Neural ODEs into two parts: a physics-informed term and an unknown discrepancy term. The framework is tested using a spring-mass model as a 4-degree-of-freedom dynamical system with cubic nonlinearity, with also noisy measured data. Furthermore, they use experimental data to learn the governing dynamics of a structure equipped with a negative stiffness device  \citep{Lai2021_StructuralIdentificationPhysics_MylLMNC}.

\cite{Zha2020_PhysicsInformedMulti_LiuZLS} employ deep long short-term memory (LSTM) networks in the PINN approach to solve nonlinear structural system subjected to seismic excitation, like steel moment resistant frame and the  single degree-of-freedom Bouc–Wen model, a nonlinear system with rate-dependent hysteresis \citep{Zha2020_PhysicsInformedMulti_LiuZLS}. 
In general they tried to address the problems of nonlinear equation of motion :
\begin{equation*}
    \ddot{\mathbf{u}} + \mathbf{g} = -\boldsymbol{\Gamma}{a}_g
\end{equation*}
where $\mathbf{g}(t)=\textbf{M}^{-1}\mathbf{h}(t)$ denotes the mass-normalized restoring force, being \textbf{M} the mass matrices; \textbf{h} the total nonlinear restoring force, and $\boldsymbol{\Gamma}$ force distribution vector.

Directed graph models can be used to directly implement ODE as deep neural networks \citep{Via2021_EstimatingModelInadequacy_NasVNDY}, while using  an Euler RNN for numerical integration.

In \cite{Nas2020_TutorialSolvingOrdinary_FriNFV} is presented a tutorial on how to use Python to implement the integration of ODEs using recurrent neural networks.

ODE-net idea is used in \cite{Ton2021_SymplecticNeuralNetworks_XioTXH} for creating Symplectic Taylor neural networks. These NNs consist of two sub-networks, that use symplectic integrators instead of Runge-Kutta, as done originally in ODE-net, which are based on residual blocks calculated with the Euler method. Hamiltonian systems as Lotka–Volterra,  Kepler, and  Hénon–Heiles systems are also tested in the aforementioned paper.

\subsection{Partial Differential Equations}
Partial Differential Equations are the building bricks of a large part of models that are used to mathematically describe physics phenomenologies. Such models have been deeply investigated and often solved with the help of different numerical strategies. Stability and convergence of these strategies have been deeply investigated in literature, providing a solid theoretical framework to approximately solve differential problems. In this Section, the application of the novel methodology of PINNs on different typologies of Partial Differential models is explored. 

\subsubsection{Steady State PDEs}
In \cite{Kha2021_HpVpinnsVariational_ZhaKZK, Kha2019_VariationalPhysicsInformed_ZhaKZK}, a general steady state problem is addressed as:
\begin{align*}
 \mathcal{F}_s(\bm{u}(\bm{x}); \bm{q}) &= f(\bm{x})  \quad \bm{x}\in \Omega ,
 \\
\mathcal{B}(\bm{u}(\bm{x})) &= 0  \quad \bm{x}\in \partial \Omega
\end{align*}
over the domain $\Omega \subset \mathbb{R}^d$ with dimensions $d$ and bounds $\partial \Omega$.  $\mathcal{F}_s$  typically contains differential and/or integro-differential operators with parameters $\bm{q}$ and $ f(\bm{x})$ indicates some forcing term.

In particular an Elliptic equation can generally be written by setting
\begin{equation*}
\mathcal{F}_s(u(x); \sigma, \mu)  = - div(\mu \nabla u)) + \sigma u
\end{equation*}

\cite{Tar2020_PhysicsInformedDeep_MarTMP}
consider a linear 
\begin{equation*}
\mathcal{F}_s(u(x); \sigma)  =   \nabla \cdot (K(\mathbf{x}) \nabla u(\mathbf{x})) = 0 
\end{equation*}
and non linear 
\begin{equation*}
\mathcal{F}_s(u(x); \sigma)  =   \nabla \cdot [K(u)\nabla u(\mathbf{x})] =0
\end{equation*}
diffusion equation with unknown diffusion coefficient $K$. The equation essentially describes an unsaturated flow in a homogeneous porous medium, where $u$ is the water pressure and $K(u)$ is the porous medium's conductivity.  It is difficult to measure $K(u)$ directly, so \cite{Tar2020_PhysicsInformedDeep_MarTMP} assume that only a finite number of measurements of $u$ are available. 

\cite{Tar2020_PhysicsInformedDeep_MarTMP} demonstrate that the PINN method outperforms the state-of-the-art maximum a posteriori probability method.
Moreover, they show that utilizing only capillary pressure data for unsaturated flow, PINNs can estimate the pressure-conductivity for unsaturated flow. 
One of the first novel approach, PINN based, was the variational physics-informed neural network (VPINN) introduced in \cite{Kha2019_VariationalPhysicsInformed_ZhaKZK},  which has the advantage of decreasing the order of the differential operator through integration-by-parts. The authors tested VPINN with the steady Burgers equation, and on the two dimensional Poisson’s equation.
VPINN \cite{Kha2019_VariationalPhysicsInformed_ZhaKZK} is also used to solve Schrödinger Hamiltonians, i.e. an elliptic reaction-diffusion operator \citep{Gru2021_DeepNeuralNetwork_HajGHL}.

In \cite{Hag2021_NonlocalPhysicsInformed_BekHBMJ} a nonlocal approach with the PINN framework is used  to solve two-dimensional quasi-static mechanics for linear-elastic and elastoplastic deformation. They define a loss function for elastoplasticity, and the input  variables to the feed-forward neural network are the displacements, while the output variables are the components of the strain tensor and stress tensor.  The localized deformation and strong gradients in the solution make the boundary value problem difficult solve.
The Peridynamic Differential Operator (PDDO) is used in a nonlocal approach with the PINN paradigm in \cite{Hag2021_NonlocalPhysicsInformed_BekHBMJ}.
They demonstrated that the PDDO framework can capture stress and strain concentrations using global functions.

In \cite{Dwi2020_PhysicsInformedExtreme_SriDS} the authors address different 1D-2D linear advection and/or diffusion steady-state problems from \cite{Ber2018_UnifiedDeepArtificial_NysBN}, by using their PIELM, a PINN combined with ELM (Extreme Learning Machine). A critical point is that the proposed PIELM only takes into account linear differential operators.

In  \cite{Ram2021_SpinnSparsePhysics_RamRR} they consider linear elliptic PDEs, such as the solution of the Poisson equation in both regular and irregular domains, by addressing non-smoothness in solutions.

The authors in \cite{Ram2021_SpinnSparsePhysics_RamRR} propose a class of partially interpretable sparse neural network architectures (SPINN), and this architecture is achieved by reinterpreting meshless representation of PDE solutions.

Laplace-Beltrami Equation is solved on 3D surfaces, like complex geometries, and high dimensional surfaces, by discussing the relationship between sample size, the structure of the PINN, and accuracy \citep{Fan2020_PhysicsInformedNeural_ZhaFZ}.

The PINN paradigm has also been applied to Eikonal equations, i.e.  are hyperbolic problems written as
\begin{equation*}
\begin{aligned}
    \| \nabla u (\bm{x})\|^2 & = \frac{1}{v^2(\bm{x})}, \qquad \forall \, \bm{x}\, \in \, \Omega,
\label{eq:eikonal}
\end{aligned}
\end{equation*}
where $v$ is a velocity and $u$ an unknown activation time. 
The Eikonal equation describes wave propagation, like the travel time of seismic wave \citep{Wah2021_PinneikEikonalSolution_HagWHA, Smi2021_EikonetSolvingEikonal_AziSAR} or cardiac activation electrical waves \citep{Sah2020_PhysicsInformedNeural_YanSCYP, Gra2021_LearningAtrialFiber_PezGPC}.

By implementing  EikoNet, for solving a 3D Eikonal equation, \cite{Smi2021_EikonetSolvingEikonal_AziSAR}  find the travel-time field in heterogeneous 3D structures; however, the proposed PINN model is only valid for a single fixed velocity model, hence changing the velocity, even slightly, requires retraining the neural network. EikoNet essentially predicts the time required to go from a source location to a receiver location, and it has a wide range of applications, like earthquake detection. 

PINN is also proved to outperform the first-order fast sweeping solution in accuracy tests \citep{Wah2021_PinneikEikonalSolution_HagWHA}, especially in the anisotropic model.

Another approach involves synthetic and patient data for learning heart tissue fiber orientations from electroanatomical maps, modeled with anisotropic Eikonal equation \citep{Gra2021_LearningAtrialFiber_PezGPC}. In their implementation the authors add to the loss function also a Total Variation regularization for the conductivity vector.

By neglecting anisotropy, cardiac activation mapping is also addressed by \cite{Sah2020_PhysicsInformedNeural_YanSCYP} where PINNs are used with randomized prior functions to quantify data uncertainty and create an adaptive sampling strategy for acquiring and creating activation maps.

Helmholtz equation for weakly inhomogeneous two-dimensional (2D) media under transverse magnetic polarization excitation
 is addressed in \cite{Che2020_PhysicsInformedNeural_LuCLK} as:
\begin{equation*}
\nabla^{2}{E_z\left(x,y\right)}+{\varepsilon_{r}}\left(x,y\right)k_{0}^{2}E_{z}=0,
\end{equation*}
whereas high frequency Helmholtz equation (frequency domain Maxwell’s equation) is solved in \cite{Fan2020_DeepPhysicalInformed_ZhaFZ}.

\subsubsection{Unsteady PDEs}
Unsteady PDEs usually describe the evolution in time of a physics phenomena. Also in this case, PINNs have proven their reliability in solving such type of problems resulting in a flexible methodology. 

\subsubsubsection{Advection-Diffusion-Reaction Problems}
Originally \cite{Rai2019_PhysicsInformedNeural_PerRPK} addressed unsteady state problem as:%
\begin{align*}
u_t &= \mathcal{F}_x(u(\bm{x})) \quad \bm{x}\in \Omega ,
 \\
\mathcal{B}(u(\bm{x})) &= 0  \quad \bm{x}\in \partial \Omega
\end{align*}
where $\mathcal{F}_x$  typically contains differential operators of the variable $x$.
In particular a general advection-diffusion reaction  equation can be written by setting
\begin{equation*}
\mathcal{F}_x(u(x);  \bm{b},\mu, \sigma )  = - div(\mu \nabla u)) + \bm{b}\nabla u + \sigma u,
\end{equation*}
where, given the parameters $\bm{b},\mu, \sigma$,  $- div(\mu \nabla u))$ is the \emph{diffusion} term, while the advection term is $\bm{b}\nabla u$ which is also known as \emph{transport} term, and finally $\sigma u$ is the \emph{reaction} term.

\paragraph{Diffusion Problems}
For a composite material, \cite{Ami2021_PhysicsInformedNeural_HagANHC} study a system of equations, that models heat transfer with the known heat equation, 
\begin{equation*} 
    \frac{\partial T}{\partial t} = 
    a \frac{\partial^2 T}{\partial x^2}
    + b \frac{d \alpha}{d t}
\end{equation*}
where $a,b$ are parameters, and a second equation for  internal heat generation  expressed as a derivative of time of the degree of cure $\alpha \in (0,1)$ is present.

\cite{Ami2021_PhysicsInformedNeural_HagANHC} propose a PINN composed of two disconnected subnetworks and the use of a sequential training algorithm that automatically adapts the weights in the loss, hence increasing the model's prediction accuracy. 

Based on physics observations, an activation function with a positive output parameter and a non-zero derivative is selected for the temperature describing network's last layer, i.e. a Softplus activation function, that is a smooth approximation to the ReLU activation function.
The Sigmoid function is instead chosen for the last layer of the network that represents the degree of cure. 
Finally, because of its smoothness and non-zero derivative, the hyperbolic-tangent function is employed as the activation function for all hidden layers. 

Since accurate exotherm forecasts are critical in the processing of composite materials inside autoclaves, \cite{Ami2021_PhysicsInformedNeural_HagANHC} show that PINN correctly predict the maximum part temperature, i.e. exotherm, that occurs in the center of the composite material due to internal heat.

A more complex problem was addressed in \cite{Cai2021_PhysicsInformedNeural_WanCWW}, where the authors study a kind of free boundary problem, known as the Stefan problem.

The Stefan problems can be divided into two types: the direct Stefan problem, in its traditional form, entails determining the temperature distribution in a domain during a phase transition. The latter, inverse problem, is distinguished by a set of free boundary conditions known as Stefan conditions \cite{Wan2021_DeepLearningFree_PerWP}.

The authors characterize temperature distributions using a PINN model, that consists of a DNN to represent the unknown interface and another FCNN with two outputs, one for each phase. This leads to three residuals, each of which is generated using three neural networks, namely the two phases $u_{\theta}^{(1)}$, $u_{\theta}^{(2)}$, as well as the interface $s_{\beta}$ that takes the boundary conditions into consideration. 
The two sets of parameters $\theta$ and $\beta$ are minimized through the mean squared errors losses:
\begin{equation*}
\mathcal{L}_\mathcal{F}(\theta) =  \mathcal{L}_r^{(1)}(\theta) +  \mathcal{L}_r^{(2)}(\theta) 
\end{equation*}
enforces the two PDEs of the heat equation, one for each phase state:
\begin{equation*}
\mathcal{L}_r^{(k)}(\theta) =  \frac{1}{N_c} \sum_{i=1}^{N_c} \|
\frac{\partial u_{\theta}^{(k)}  }{\partial t} (x^i,t^i)
- \omega_k \frac{\partial^2 u_{\theta}^{(k)}}{\partial x^2}  (x^i,t^i)
\|^2, \qquad k=1,2.
\end{equation*}
on a set of randomly selected collocation locations $\{ (x^i,t^i)\}_{i=1}^{N_c}$,  and $\omega_1, \omega_2$ are two additional training parameters.
While, as for the boundary and initial conditions:
\begin{equation*}
\mathcal{L}_\mathcal{B}(\theta)  = 
\mathcal{L}_{s_{bc}}^{(1)}(\theta,\beta) +  \mathcal{L}_{s_{bc}}^{(2)}(\theta,\beta)+
\mathcal{L}_{s_{Nc}}(\theta,\beta) +  \mathcal{L}_{s_{0}}(\beta)
\end{equation*}
where $\mathcal{L}_{s_{bc}}^{(k)}$ are the boundary condition of $u^{(k)}$ on the moving boundary $s(t)$, $\mathcal{L}_{s_{Nc}}$ is the free boundary Stefan problem equation, and $\mathcal{L}_{s_{0}}$ is the initial condition on the free boundary function.
Finally, as for the data,
\begin{equation*}
\mathcal{L}_{data}(\theta)  = 
\frac{1}{N_d}\sum\limits_{i=1}^{N_d}
\|u_{\theta}(x^i_{data}, t^i_{data}) - u_i^*\|^2 ,
\end{equation*}
computes the error of the approximation $u(x,t)$ at known data points.

With the previous setup the authors in \cite{Cai2021_PhysicsInformedNeural_WanCWW} find an accurate solution, however, the basic PINN model fails to appropriately identify the unknown thermal diffusive values, for the inverse problem,  due to a local minimum in the training procedure.

So they employ a dynamic weights technique \citep{Wan2021_UnderstandingMitigatingGradient_TenWTP}, which mitigates problems that arise during the training of PINNs due to stiffness in the gradient flow dynamics.
The method significantly minimizes the relative prediction error, showing that the weights in the loss function are crucial and that choosing ideal weight coefficients can increase the performance of PINNs \citep{Cai2021_PhysicsInformedNeural_WanCWW}.

In \cite{Wan2021_DeepLearningFree_PerWP}, the authors conclude that the PINNs prove to be versatile in approximating complicated functions like the Stefan problem, despite the absence of adequate theoretical analysis like approximation error or numerical stability. 

\paragraph{Advection Problems}
In \cite{He2021_PhysicsInformedNeural_TarHT}, multiple advection-dispersion equations are addressed, like
\begin{equation*}
u_t + \nabla \cdot (-\kappa \nabla u + \vec{v} u)  = s
\end{equation*}
where $\kappa$ is the dispersion coefficient.

The authors find that the PINN method is accurate and produces results that are superior to those obtained using typical discretization-based methods. 
Moreover both \cite{Dwi2020_PhysicsInformedExtreme_SriDS} and \cite{He2021_PhysicsInformedNeural_TarHT}  solve the same 2D advection-dispersion equation,
\begin{equation*}
u_t + \nabla \cdot (-\kappa \nabla u + \vec{a} u) = 0
\end{equation*}
In this comparison, the PINN technique \citep{He2021_PhysicsInformedNeural_TarHT} performs better that the ELM method \citep{Dwi2020_PhysicsInformedExtreme_SriDS}, given the errors that emerge along borders, probably due to larger wights assigned to boundary and initial conditions in \cite{He2021_PhysicsInformedNeural_TarHT}.

Moreover, in \cite{Dwi2020_PhysicsInformedExtreme_SriDS}, an interesting case of PINN and PIELM failure in solving the linear advection equation is shown, involving PDE with sharp gradient solutions.

\cite{He2020_PhysicsInformedNeural_BarHBTT} solve Darcy and advection–dispersion equations proposing a Multiphysics-informed neural network  (MPINN) for subsurface transport problems, and also explore the influence of the neural network size on the accuracy of parameter and state estimates.

In \cite{Sch2021_ExtremeTheoryFunctional_FurSFL}, a comparison of two methods is shown, Deep-TFC and X-TFC, on how the former performs better in terms of accuracy when the problem becomes sufficiently stiff. The examples are mainly based on  1D time-dependent Burgers’ equation and the Navier–-Stokes (NS) equations.

In the example of the two-dimensional Burgers equation, \cite{Jag2020_ConservativePhysicsInformed_KhaJKK} demonstrate that by having an approximate a priori knowledge of the position of shock, one can appropriately partition the domain to capture the steep descents in solution. 
This is accomplished through the cPINN domain decomposition flexibility. 

While \cite{Arn2021_StateModelingControl_KinAK} addressed the Burgers equations in the context of model predictive control (MPC). 

In \cite{Men2020_PpinnPararealPhysics_LiMLZK} the authors study a two-dimensional diffusion-reaction equation that involves long-time integration and they use a parareal PINN (PPINN) to divide the time interval into equal-length sub-domains.
PPINN is composed of a fast coarse-grained (CG) solver and a finer solver given by PINN.

\subsubsubsection{Flow Problems}\\
Particular cases for unsteady differential problems are the ones connected to the motion of fluids. Navier-Stokes equations are widely present in literature, and connected to a large number of problems and disciplines. This outlines the importance that reliable strategies for solving them has for the scientific community. Many numerical strategies have been developed to solve this kind of problems. However, computational issues connected to specific methods, as well as difficulties that arise in the choice of the discrete spatio-temporal domain may affect the quality of numerical solution. PINNs, providing mesh-free solvers, may allow to overcome some issues of standard numerical methods, offering a novel perspective in this field.  

\paragraph{Navier-Stokes Equations}
Generally Navier-Stokes equations are written as
\begin{equation*}
\mathcal{F}_x(u(x);  \nu , p )  = - div[\nu (\nabla u+ \nabla u^T)] + (u+\nabla) u +\nabla p -\bm{f},
\end{equation*}
where, $u$ is the speed of the fluid, $p$ the pressure and $ \nu$ the viscosity \citep{Qua2013_NumericalModelsDifferential_Qua}.
The dynamic equation is coupled with
\begin{equation*}
div( u ) = 0 
\end{equation*}
for expressing mass conservation.

The Burgers equation, a special case of the Navier-Stokes equations, was covered in the previous section.

Using quick parameter sweeps, \cite{Art2021_ActiveTrainingPhysics_KinAK} demonstrate how PINNs may be utilized to determine the degree of narrowing in a tube. PINNs are trained using finite element data to estimate Navier-Stokes pressure and velocity fields throughout a parametric domain.
The authors present an active learning algorithm (ALA) for training PINNs to predict PDE solutions over vast areas of parameter space by combining ALA, a domain and mesh generator, and a traditional PDE solver with PINN.

PINNs are also applied on the drift-reduced Braginskii model by learning turbulent fields using limited electron pressure data \citep{Mat2021_UncoveringTurbulentPlasma_FraMFH}. The authors simulated synthetic plasma using the global drift-ballooning (GDB) finite-difference algorithm by solving a fluid model, ie.  two-fluid drift-reduced Braginskii equations. They also observe the possibility to infer 3D turbulent fields from only 2D observations and representations of the evolution equations. This can be used for fluctuations that are difficult to monitor or when plasma diagnostics are unavailable.

\cite{Xia2020_FlowsOverPeriodic_WuXWLD} review available turbulent flow databases and propose  benchmark datasets by systematically altering flow conditions. 

\cite{Zhu2021_MachineLearningMetal_LiuZLY} predict the temperature and melt pool fluid dynamics in 3D metal additive manufacturing AM processes.

The thermal-fluid model is characterized by Navier-Stokes equations (momentum and mass conservation), and energy conservation equations.

They approach the Dirichlet BC in a ``hard'' manner, employing a specific piece of the neural network to solely meet the prescribed Dirichlet BC; while Neumann BCs, that account for surface tension, are treated conventionally by adding the term to the loss function. They choose the loss weights based on the ratios of the distinct components of the loss function \citep{Zhu2021_MachineLearningMetal_LiuZLY}.

\cite{Che2021_DeepLearningMethod_ZhaCZ} solve fluid flows dynamics with Res-PINN, PINN paired with a Resnet blocks, that is used to improve the stability of the neural network. They validate the model with Burgers' equation and Navier-Stokes (N-S) equation, in particular, they deal with the cavity flow and flow past cylinder problems.
A curious phenomena observed by \cite{Che2021_DeepLearningMethod_ZhaCZ} is a difference in magnitude between the predicted and actual pressure despite the fact that the distribution of the pressure filed is essentially the same. 

To estimate the solutions of parametric Navier–Stokes equations, \cite{Sun2020_SurrogateModelingFluid_GaoSGPW} created a physics-constrained, data-free, FC-NN  for incompressible flows. The DNN is trained purely by reducing the residuals of the governing N-S conservation equations, without employing CFD simulated data. 
The boundary conditions are also hard-coded into the DNN architecture, since the aforementioned authors claim  that in data-free settings, ``hard'' boundary enforcement is preferable than ``soft'' boundary approach. 

Three flow examples relevant to cardiovascular applications were used to evaluate the suggested approaches. 

In particular, the Navier--Stokes equations are given \citep{Sun2020_SurrogateModelingFluid_GaoSGPW} as: 
\begin{equation*}
	\mathcal{F}(\mathbf{u}, p) = 0 := \left \{
	\begin{aligned}
	&\nabla \cdot \mathbf{u} = 0,  &\mathbf{x}, t \in \Omega, \boldsymbol{\gamma} \in \mathbb{R}^d,\\
	&\frac{\partial\mathbf{u}}{\partial t} + (\mathbf{u}\cdot\nabla)\mathbf{u} + \frac{1}{\rho}\nabla p - \nu\nabla^2\mathbf{u} + \mathbf{b}_f = 0, &\mathbf{x}, t \in  \Omega, \boldsymbol{\gamma} \in \mathbb{R}^d 
	\end{aligned} \right .
\end{equation*}
where $\boldsymbol{\gamma}$ is a parameter vector, and with
\begin{equation*}
	\begin{aligned}
	\mathcal{I}(\mathbf{x}, p, \mathbf{u}, \boldsymbol{\gamma}) &= 0, \qquad & &\mathbf{x} \in \Omega, t =0,\boldsymbol{\gamma} \in \mathbb{R}^d,\\
	\mathcal{B}(t, \mathbf{x}, p, \mathbf{u}, \boldsymbol{\gamma}) &= 0, \qquad & &\mathbf{x}, t \in \partial\Omega \times [0, T], \boldsymbol{\gamma} \in \mathbb{R}^d,
	\end{aligned}
\end{equation*}
where $\mathcal{I}$ and $\mathcal{B}$ are generic differential operators that determine the initial and boundary conditions.

The boundary conditions (IC/BC) are addressed individually in \cite{Sun2020_SurrogateModelingFluid_GaoSGPW}. The Neumann BC are formulated into the equation loss, i.e., in a soft manner, whereas the IC and Dirichlet BC are encoded in the DNN, i.e., in a hard manner.

As a last example, NSFnets \citep{Jin2021_NsfnetsNavierStokes_CaiJCLK} has been developed considering two alternative mathematical representations of the Navier--Stokes equations: the velocity-pressure (VP) formulation and the vorticity-velocity (VV) formulation.

\paragraph{Hyperbolic Equations}
Hyperbolic conservation law is used to simplify the  Navier–Stokes equations in hemodynamics \citep{Kis2020_MachineLearningCardiovascular_YanKYH}. 

Hyperbolic partial differential equations are also addressed by \cite{Abr2021_StudyFeedforwardNeural_FloAF}: in particular, they study the inviscid nonlinear Burgers' equation and 1D Buckley-Leverett two-phase problem. They actually try to address problems of the following type:
\begin{equation*}
\frac{\partial u}{\partial t}+
\frac{\partial H(u)}{\partial x} = 0, 
\quad x \in \mathbb{R}, \quad t > 0, \quad \quad \quad u(x,0) = u_0(x),
\label{ivp2b}
\end{equation*}
whose results were compared with those obtained by the Lagrangian-Eulerian and Lax-Friedrichs schemes.
While \cite{Pat2022_ThermodynamicallyConsistentPhysics_ManPMT} proposes a PINN for discovering thermodynamically consistent equations that ensure hyperbolicity for inverse problems in shock hydrodynamics.
\\

Euler equations are hyperbolic conservation laws that might permit discontinuous solutions such as shock and contact waves, and in particular a one dimensional Euler system is written as \citep{Jag2020_ConservativePhysicsInformed_KhaJKK}
\begin{equation*}
   \frac{\partial U}{\partial t} + \nabla\cdot {f}(U) = 0, \; x\in \Omega\subset \mathbb{R}^2,
\end{equation*}
where
\begin{equation*}
U = \begin{bmatrix}
           \rho \\
           \rho u \\
           \rho E
         \end{bmatrix}
\qquad
f = \begin{bmatrix}
           \rho u, \\
           p+ \rho u^2 \\
           p u + \rho u E
         \end{bmatrix}
\end{equation*}
given $\rho$ as the density, $p$ as the pressure,  $u$  the velocity, and $E$ the total energy. These equations regulate a variety of high-speed fluid flows, including transonic, supersonic, and hypersonic flows. \cite{Mao2020_PhysicsInformedNeural_JagMJK} can precisely capture such discontinuous flows solutions for one-dimensional Euler equations, which is a challenging task for existing numerical techniques. 
According to \cite{Mao2020_PhysicsInformedNeural_JagMJK}, appropriate clustering of training data points around a high gradient area can improve solution accuracy in that area and reduces error propagation to the entire domain. This enhancement suggests the use of a separate localized powerful network in the region with high gradient solution, resulting in the development of a collection of individual local PINNs with varied features that comply with the known prior knowledge of the solution in each sub-domain. 
As done in \cite{Jag2020_ConservativePhysicsInformed_KhaJKK}, cPINN splits  the domain into a number of small subdomains in which multiple neural networks with different architectures (known as sub-PINN networks) can be used to solve the same underlying PDE. 

Still in reference to \cite{Mao2020_PhysicsInformedNeural_JagMJK}, the authors solve the one-dimensional Euler equations and a two-dimensional oblique shock wave problem. The authors can capture the solutions with only a few scattered points distributed randomly around the discontinuities.
The above-mentioned paper employ density gradient and pressure $p(x, t)$ data, as well as conservation laws, to infer all states of interest (density, velocity, and pressure fields) for the inverse problem without employing any IC/BCs. They were inspired by the experimental photography technique of Schlieren.

They also illustrate that the position of the training points is essential for the training process.
The results produced by combining the data with the Euler equations in characteristic form outperform the results obtained using conservative forms. 

\subsubsubsection{Quantum Problems}
A 1D nonlinear Schr\"{o}dinger equation is addressed in  \cite{Rai2018_DeepHiddenPhysics_Rai}, and  \cite{Rai2019_PhysicsInformedNeural_PerRPK} as:
\begin{equation}\label{eq:Schrodinger}
    i\frac{\partial \psi}{\partial t} + 
    \frac{1}{2} \frac{\partial^2 \psi}{\partial x^2}
    + \lvert\psi\rvert^2 \psi = 0
\end{equation}
where $\psi(x,t)$ is the complex-valued solution.
This problem was chosen to demonstrate the PINN's capacity to handle complex-valued answers and we will develop this example in~section \ref{sec:Schr_example}.
Also a quantum harmonic oscillator (QHO) 
\begin{equation*}
i \frac{\partial \psi (\bm{x},t)}{\partial t} 
+
\frac{1}{2}\varDelta \psi (\bm{x},t) - V(\bm{x},t) = 0
\end{equation*}
is addressed in \cite{Sti2020_LargeScaleNeural_BetSBB}, with $V$ a scalar potential.
They propose a gating network that determines which MLP to use, while each MLP consists of linear layers and $\tanh$ activation functions; so the solution becomes a weighted sum of MLP predictions. 
The quality of the approximated solution of PINNs was comparable to that of state-of-the-art spectral solvers that exploit domain knowledge by solving the equation in the Fourier domain or by employing Hermite polynomials. 

Vector solitons, which are solitary waves with multiple components in the the coupled nonlinear Schrödinger equation (CNLSE) is addressed by
\cite{Mo2022_DataDrivenVector_LinMLZ}, who extended PINNs with a pre-fixed multi-stage training algorithm. These findings can be extendable to similar type of equations, such as equations with a rogue wave \citep{Wan2021_DataDrivenRogue_YanWY} solution or the Sasa-Satsuma equation and Camassa-Holm equation.

\subsection{Other Problems}
Physics Informed Neural Networks have been also applied to a wide variety of problems that go beyond classical differential problems. As examples, in the following, the application of such a strategy as been discussed regarding Fractional PDEs and Uncertainty estimation. 

\subsubsection{Differential Equations of Fractional Order}
Fractional PDEs can be used to model a wide variety of phenomenological properties found in nature with parameters that must be calculated from experiment data, however in the spatiotemporal domain, field or experimental measurements are typically scarce and may be affected by noise \citep{Pan2019_FpinnsFractionalPhysics_LuPLK}. 
Because automatic differentiation is not applicable to fractional operators, the construction of PINNs for fractional models is more complex. One possible solution is to calculate the fractional derivative using the L1 scheme \citep{Meh2019_DiscoveringUniversalVariable_PanMPSK}.

In the first example from \cite{Meh2019_DiscoveringUniversalVariable_PanMPSK} they solve a turbulent flow with one dimensional mean flow.

In \cite{Pan2019_FpinnsFractionalPhysics_LuPLK}, the authors focus on identifying the parameters of fractional PDEs with known overall form but unknown coefficients and unknown operators, by giving rise to fPINN.
They construct the loss function using a hybrid technique that includes both automatic differentiation for integer-order operators and numerical discretization for fractional operators. 
They also analyse the convergence of fractional advection-diffusion equations (fractional ADEs)

The solution proposed in \cite{Pan2019_FpinnsFractionalPhysics_LuPLK}, is then extended in \cite{Kha2021_IdentifiabilityPredictabilityInteger_CaiKCZ}
where they address also time-dependent fractional orders.
The formulation in \cite{Kha2021_IdentifiabilityPredictabilityInteger_CaiKCZ} uses separate neural network to represent each fractional order and use a large neural network to represent states. 

\subsubsection{Uncertainty Estimation}
In data-driven PDE solvers, there are several causes of uncertainty.
The quality of the training data has a significant impact on the solution's accuracy. 

To address forward and inverse nonlinear problems represented by partial differential equations (PDEs) with noisy data, \cite{Yan2021_BPinnsBayesian_MenYMK}  propose a Bayesian physics-informed neural network (B-PINN).
The Bayesian neural network acts as the prior in this Bayesian framework, while an Hamiltonian Monte Carlo (HMC) method or variational inference (VI) method can be used to estimate the posterior.

B-PINNs \citep{Yan2021_BPinnsBayesian_MenYMK} leverage both physical principles and scattered noisy observations to give predictions and quantify the aleatoric uncertainty coming from the noisy data.

\cite{Yan2021_BPinnsBayesian_MenYMK} test their network on some forward problems (1D Poisson equation, Flow in one dimension across porous material with a boundary layer, Nonlinear Poisson equation in one dimension and the 2D Allen-Cahn equation), while for inverse problems the 1D diffusion-reaction system with nonlinear source term and 2D nonlinear diffusion-reaction system are addressed.

\cite{Yan2021_BPinnsBayesian_MenYMK} use also the B-PINNs for a high-dimensional diffusion-reaction system, where the locations of three contaminating sources are inferred from a set of noisy data.

\cite{Yan2020_PhysicsInformedGenerative_ZhaYZK} considers the solution of elliptic stochastic differential equations (SDEs) that required approximations of three stochastic processes: the solution $u(x;\gamma)$, the forcing term $f(x;\gamma)$, and the diffusion coefficient $k(x;\gamma)$. 

In particular, \cite{Yan2020_PhysicsInformedGenerative_ZhaYZK}  investigates the following, time independent, SDE
\begin{equation*}
\begin{aligned}
\mathcal{F}_{\boldsymbol{x}}[u(\boldsymbol{x} ; \gamma) ; k(\boldsymbol{x} ; \gamma)] &=f(\boldsymbol{x} ; \gamma),
\\
B_{\boldsymbol{x}}[u(\boldsymbol{x} ; \gamma)] &=b(\boldsymbol{x} ; \gamma), 
\end{aligned}
\end{equation*}
where $k(x; \gamma )$ and $f (x; \gamma )$ are independent stochastic processes, with $k$ strictly positive.

Furthermore, they investigate what happens when there are a limited number of measurements from scattered sensors for the stochastic processes.
They show how the problem gradually transform from forward to mixed, and finally to inverse problem. This is accomplished by assuming that there are a sufficient number of measurements from some sensors for $f(x;\gamma)$, and then as the number of sensors measurements for $k(x;\gamma)$ is decreased, the number of sensors measurements on $u(x;\gamma)$ is increased, and thus a forward problem is obtained when there are only sensors for $k$ and not for $u$, while an inverse problem has to be solved when there are only sensors for $u$ and not for $k$.

In order to characterizing shape changes (morphodynamics) for cell-drug interactions, \cite{Cav2021_PhysicsInformedDeep_MosCMS} use kernel density estimation (KDE) for translating morphspace embeddings into probability density functions (PDFs). Then they use  a top-down Fokker-Planck model of diffusive development over Waddington-type landscapes, with a PINN learning such landscapes by fitting the PDFs to the Fokker–Planck equation.
The architecture includes a neural network for each condition to learn: the PDF, diffusivity, and landscape.
All parameters are fitted using approximate Bayesian computing with sequential Monte Carlo (aBc-SMC) methods: in this case, aBc selects parameters from a prior distribution and runs simulations; if the simulations match the data within a certain level of similarity, the parameters are saved.
So the posterior distribution is formed by the density over the stored parameters  \citep{Cav2021_PhysicsInformedDeep_MosCMS}.

\subsection{Solving a differential problem with PINN}\label{sec:Schr_example}

\begin{figure}[hbt!]
     \centering
     \begin{subfigure}[b]{0.49\textwidth}
         \centering
         \includegraphics[width=\textwidth]{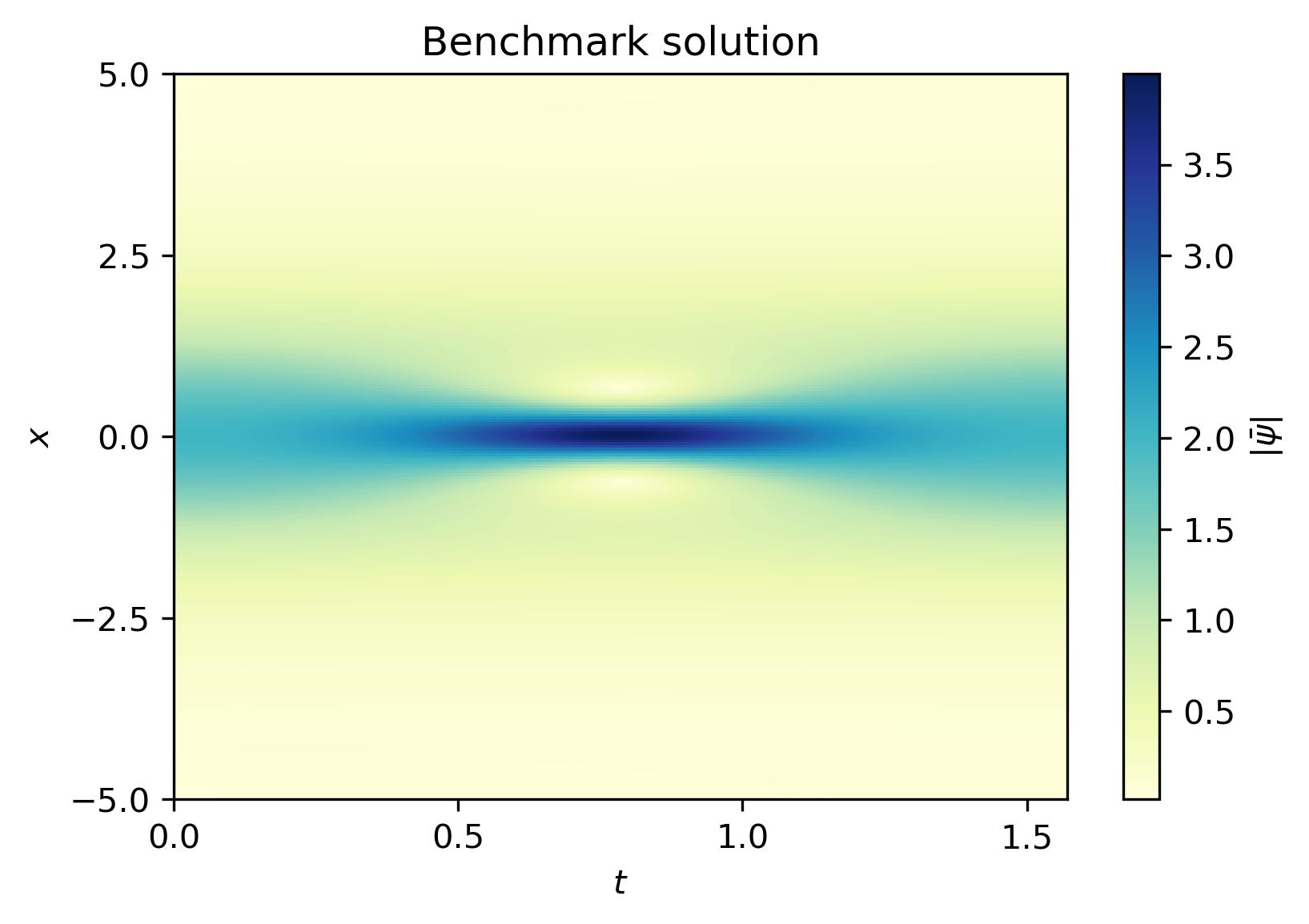}
     \end{subfigure}
     \hfill
     \begin{subfigure}[b]{0.49\textwidth}
         \centering
         \includegraphics[width=\textwidth]{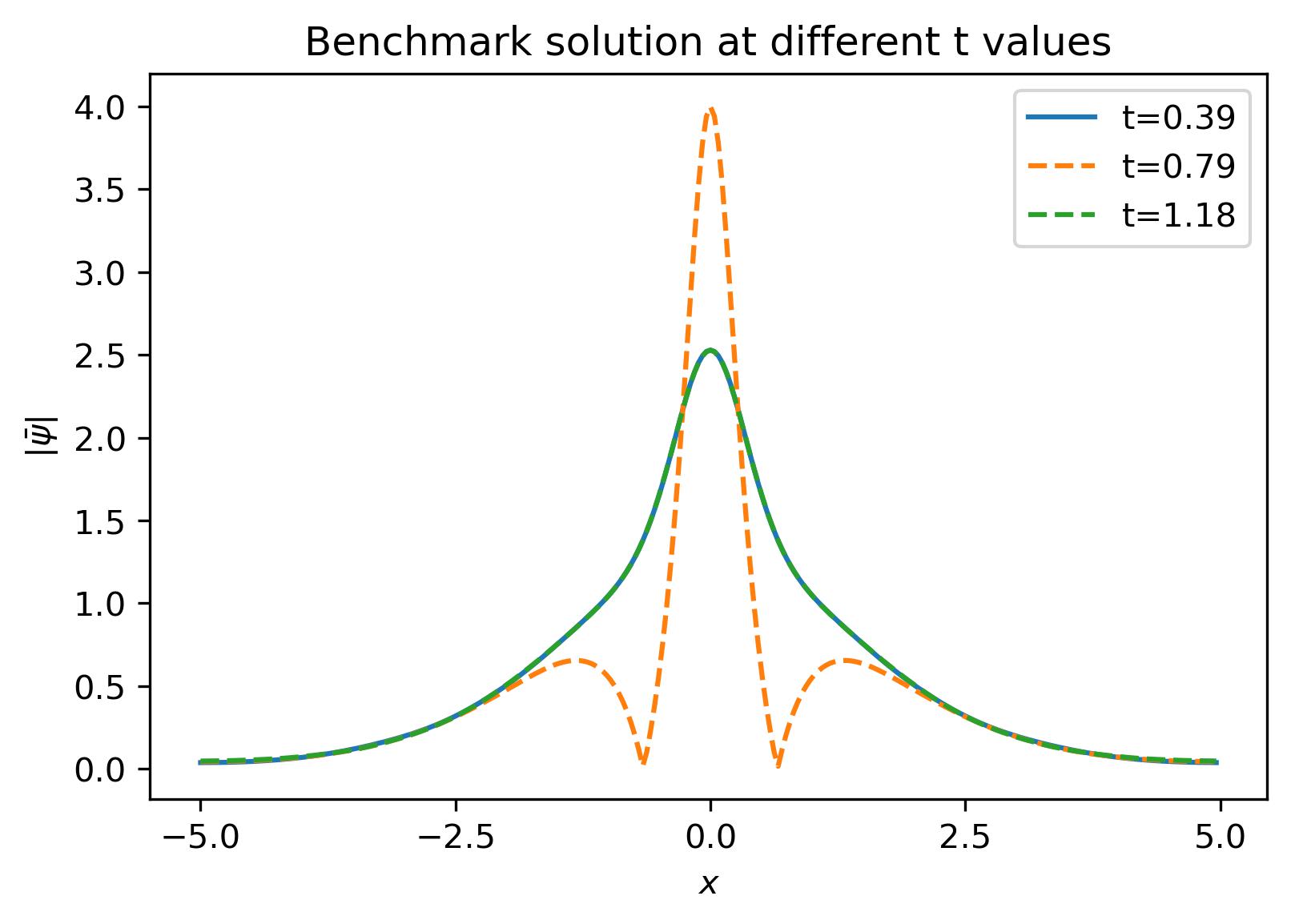}
     \end{subfigure}
        \caption{1D nonlinear Shr\"{o}dinger (NLS) solution module $\lvert\bar{\psi}\rvert$. The solution is generated with Chebfun open-source software \citep{Driscoll2014} and used as a reference solution. We also show the solution at three different time frames, $t=\frac{\pi}{8}, \frac{\pi}{4}, \frac{3\pi}{8}$. We expect the solution to be symmetric with respect to $\frac{\pi}{4}$. }
        \label{fig:Chebfun_NLS}
\end{figure}

Finally, this subsection discusses a realistic example of a 1D nonlinear Shr\"{o}dinger (NLS) problem, as seen in Figure~\ref{fig:Chebfun_NLS}.
The nonlinear problem is the same presented in \cite{Rai2018_DeepHiddenPhysics_Rai, Rai2017_PhysicsInformedDeep1_PerRPK}, used to demonstrate the PINN’s ability to deal with periodic boundary conditions and complex-valued solutions.

Starting from an initial state $\psi(x,0) = 2\ \text{sech}(x)$ and assuming periodic boundary conditions equation~\eqref{eq:Schrodinger} with all boundary conditions results in the initial boundary value problem, given a domain $\Omega = [-5, 5] \times (0,T]$  written as:
\begin{equation}\label{eq:IBVPSchrodinger}
  \begin{cases}
    {\begin{alignedat}{3}
    & i \psi_t + 0.5 \psi_{xx} + \lvert\psi\rvert^2 \psi = 0   & \qquad & (x,t)\in \Omega  \\
     & \psi(0,x) = 2\ \text{sech}(x) &&  x \in[-5, 5] \\
     & \psi(t,-5) = \psi(t,5)
     &&  t \in (0,T]\\
     & \psi_x(t,-5) = \psi_x(t,5)
     &&  t \in (0,T]
    \end{alignedat}}
  \end{cases}
\end{equation}
where 
$T=\pi/2$.

To assess the PINN’s accuracy, \cite{Rai2017_PhysicsInformedDeep1_PerRPK} created a high-resolution data set by simulating the Schrödinger equation using conventional spectral methods.
The authors integrated the Schrödinger equation up to a final time $T=\pi/2$ using the MATLAB based Chebfun open-source software
\citep{Driscoll2014}.

The PINN solutions are trained on a subset of measurements, which includes initial data, boundary data, and collocation points inside the domain. 
The initial time data, $t=0$, are  $\{x_0^i, \psi^i_0\}_{i=1}^{N_0}$, the boundary collocation points are $\{t^i_b\}_{i=1}^{N_b}$, and the collocation points on $\mathcal{F}(t,x)$ are $\{t_c^i,x_c^i\}_{i=1}^{N_c}$.
In \cite{Rai2017_PhysicsInformedDeep1_PerRPK}
a total of $N_0 = 50$ initial data points on $\psi(x,0)$  are randomly sampled from the whole high-resolution data-set to be included in the training set, as well as $N_b = 50$ randomly sampled boundary points to enforce the periodic boundaries.
Finally for the solution domain, it is assumed $N_c=20\ 000$ randomly sampled collocation points.

The neural network architecture has two inputs, one for the time $t$ and the other one the location $x$, while the output has also length 2 rather than 1, as it would normally be assumed, because the output of this NN is expected to find the real and imaginary parts of the solution.


The network is trained in order to minimize the losses due to the initial and boundary conditions, $\mathcal{L}_\mathcal{B}$,
as well as to satisfy the Schrodinger equation on the collocation points, i.e.  $\mathcal{L}_\mathcal{F}$.
Because we are interested in a model that is a surrogate for the PDE, no extra data, $\mathcal{L}_{data}=0$, is used.
In fact we only train the PINN with the known data point from the initial time step $t=0$.
So the losses of \eqref{eq:loss_pinn} are:
\begin{multline*}
\mathcal{L}_\mathcal{B}
= \frac{1}{N_0}\sum_{i=1}^{N_0} \lvert \psi(0,x_0^i) - \psi^i_0\rvert^2
+ \\
\frac{1}{N_b}\sum_{i=1}^{N_b} \left( \lvert \psi^i(t^i_b,-5) - \psi^i(t^i_b,5)\rvert^2 + \lvert \psi^i_x(t^i_b,-5) - \psi^i_x(t^i_b,5)\rvert^2 \right)
\end{multline*}
and
$$
\mathcal{L}_\mathcal{F} = \frac{1}{N_c}\sum_{i=1}^{N_c}\lvert \mathcal{F}(t_c^i,x_c^i)\rvert^2.
$$
The Latin Hypercube Sampling technique 
\citep{Stein1987LargeSP}
is used to create all randomly sampled points among the benchmark data prior to training the NN.

In our training we use Adam, with a learning rate of $10^{-3}$, followed by a final  fine-tuning with LBFGS.
We then explore different settings and architectures as in Table~\ref{tab:NLS},
by analysing the  Mean Absolute Error (MAE) and Mean Squared Error (MSE).
We used  the PyTorch implementation from \cite{Sti2020_LargeScaleNeural_BetSBB} which is accessible on GitHub. While the benchmark solutions are from the GitHub of \cite{Rai2017_PhysicsInformedDeep1_PerRPK}.

\cite{Rai2017_PhysicsInformedDeep1_PerRPK}
first used a DNN with  5 layers each with 100 neurons per layer and a hyperbolic tangent activation function in order to represent the unknown function $\psi$ for both real and imaginary parts.
In our test, we report the original configuration but we also analyze other network architectures and training point amounts.

\begin{figure}[hbt!]
     \centering
     \begin{subfigure}[b]{0.49\textwidth}
         \centering
         \includegraphics[width=\textwidth]{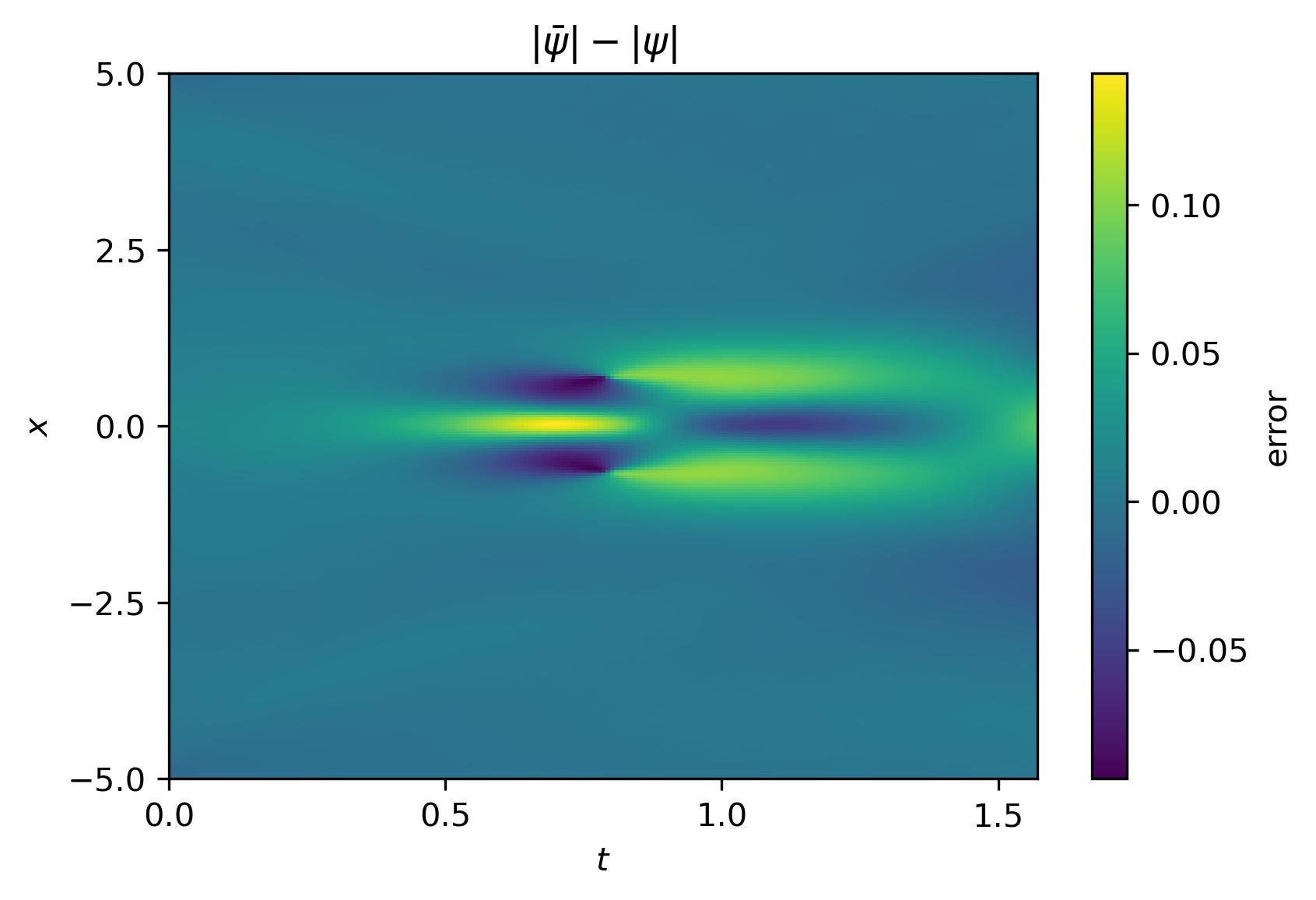}
     \end{subfigure}
     \hfill
     \begin{subfigure}[b]{0.49\textwidth}
         \centering
         \includegraphics[width=\textwidth]{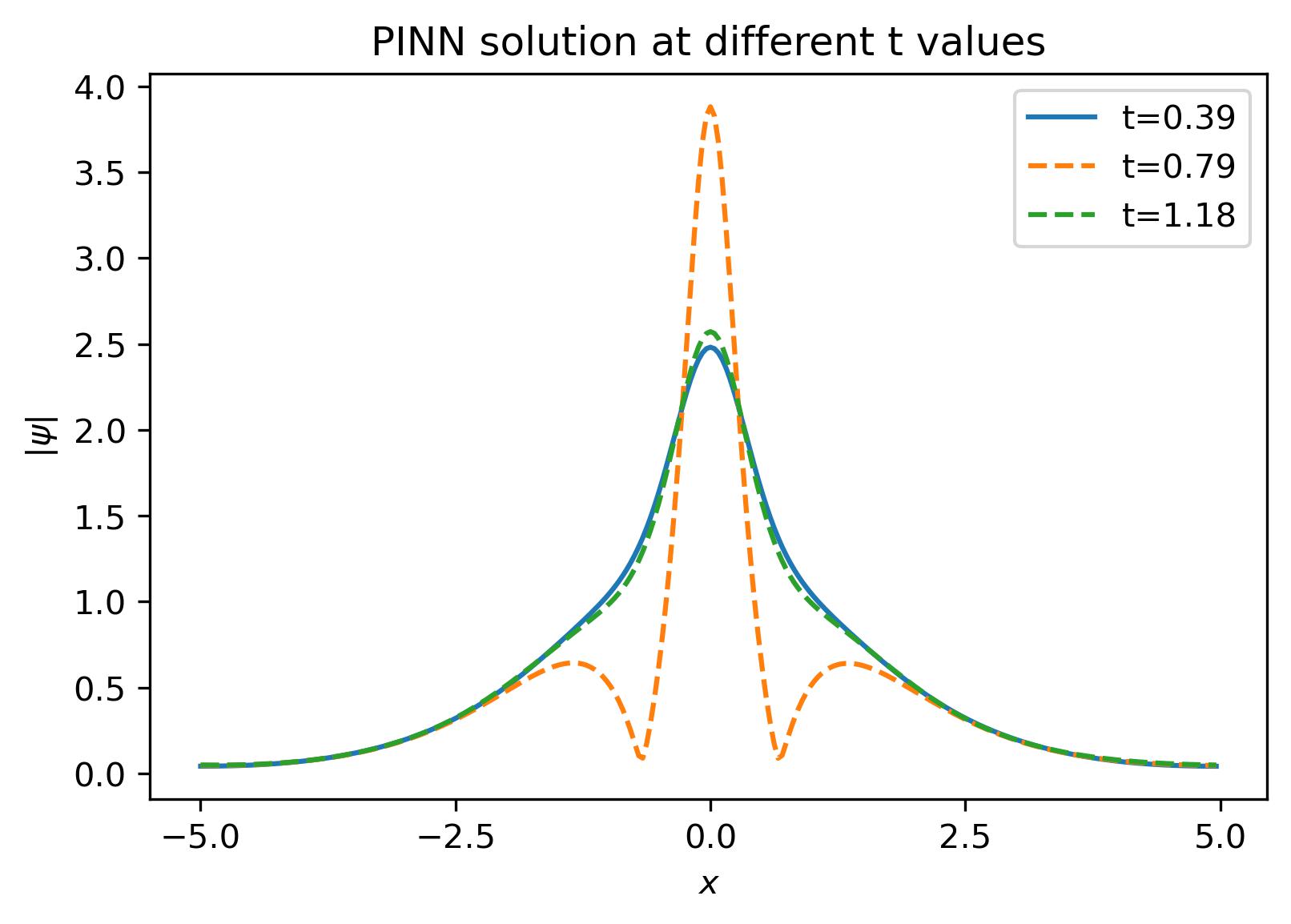}
     \end{subfigure}
        \caption{PINN solutions of the 1D nonlinear Shr\"{o}dinger (NLS) problem.
        As illustrated in the first image, the error is represented as the difference among the benchmark solution and PINN result, whereas the second image depicts the PINN's solution at three independent time steps: $t=\pi/8, \pi/4 , 3/8\pi$.
        In this case, we would have expected the solution to overlap at both $\pi/8$ and $ 3/8\pi$, but there isn't a perfect adherence; in fact, the MSE is $5,17\cdot 10^{-04}$ in this case, with some local peak error of $10^{-01}$ in the second half of the domain.
        }
        \label{fig:PINN_best_sol}
\end{figure}

A comparison of the predicted and exact solutions at three different temporal snapshots is shown in Figures~\ref{fig:Chebfun_NLS}, and~\ref{fig:PINN_best_sol}.
All the different configurations reported in Table~\ref{tab:NLS} show similar patterns of Figure~\ref{fig:PINN_best_sol}, the only difference is in the order of magnitude of the error. 

In particular in such Figure, we show the best configuration obtained in our test, with an average value of
$5,17\cdot 10^{-04}$ 
according to the MSE.
Figure~\ref{fig:PINN_best_sol} first shows the modulus of the predicted spatio-temporal solution $\lvert  \psi(x,t)\rvert$, with respect to the benchmark solution, by plotting the error at each point.
This PINN has some difficulty predicting the central height around $(x,t)=(0,\pi/4)$, as well as in mapping value on in  $t \in (\pi/4, \pi/2)$ that are symmetric to the time interval $t \in (0, \pi/4)$.

Finally, in Table~\ref{tab:NLS}, we present the results of a few runs in which we analyze what happens to the training loss, relative $L_2$, MAE, and MSE when we vary the boundary data, and initial value data.
In addition, we can examine how the error grows as the solution progresses through time.

\begin{table}[hbt!]
\centering
\begin{subtable}[t]{1.01\textwidth}
\begin{tabular}{ccccccc}
\toprule
\textbf{$N_0$} & \textbf{$N_b$} &  \textbf{$N_c$} & \textbf{Training loss} & \textbf{relative $L_2$} & \textbf{MAE} & \textbf{MSE} \\
\midrule
40 & 50 & 20000  & $9\times 10^{-4}$ & $0.025 \pm 0.003$ & $0.065 \pm 0.004$ & $0.019 \pm 0.002$ \\
40 & 100 & 20000  & $1\times 10^{-3}$ & $0.024 \pm 0.002$ & $0.065 \pm 0.003$ & $0.019 \pm 0.002 $ \\
80 & 50 & 20000  & $6 \times 10^{-4}$ & $0.007 \pm 0.001$ & $0.035 \pm 0.004$ & $0.005 \pm 0.001$ \\
80 & 100 & 20000 & $6\times 10^{-4}$ & $0.006 \pm 0.002$ & $0.033 \pm 0.005$ & $0.005 \pm 0.002$ \\
\bottomrule
\end{tabular}
\caption{Case where the NN is fixed with 100 neurons per layer and four hidden layers.
Only the number the number of training points on boundary conditions are doubled. 
}
\label{tab:table1_a}
\end{subtable}
\hspace{\fill}
\begin{subtable}[t]{0.99\textwidth}
\begin{tabular}{cccc}
\toprule
\textbf{time $t$} & \textbf{relative $L_2$} & \textbf{MAE} & \textbf{MSE} \\
\midrule                                      
0   & $(5\pm1)\times 10^{-4} $ & $0.012 \pm 0.002$ & $(4\pm 1)\times 10^{-4}$ \\
0.39   & $(15\pm 5)\times 10^{-4}$ & $0.015 \pm 0.002$ & $(12\pm 4)\times 10^{-3}$  \\
0.79  & $0.009 \pm 0.003$ & $0.038 \pm 0.006$ & $0.007 \pm 0.003$  \\
1.18  & $0.01 \pm 0.004$ & $0.051 \pm 0.009$ & $0.009 \pm 0.003$   \\
1.56  & $0.005 \pm 0.001$ & $0.044 \pm 0.005$ & $0.004 \pm 0.001$  \\
\bottomrule
\end{tabular}
\caption{Error behavior at various time steps for the best occurrence in Table, when \subref{tab:table1_a},  where $N_0 = 80$, $N_b=100$, $N_c=20000$.}
\label{tab:table1_b}
\end{subtable}
\caption{
Two subtables exploring the effect of the amount of data points on convergence for PINN and the inherent problems of vanilla pINN to adhere to the solution for longer time intervals, in solving the 1D nonlinear Shr\"{o}dinger (NLS) equation~\eqref{eq:IBVPSchrodinger}.
In \subref{tab:table1_a}, the NN consists of four layers, each of which contains $100$ neurons, and we can observe how increasing the number of training points over initial or boundary conditions will decrease error rates.
Furthermore, doubling the initial condition points has a much more meaningful influence impact than doubling the points on the spatial daily domain in this problem setup. 
In  \subref{tab:table1_b}, diven the best NN, we can observe that a vanilla PINN has difficulties in maintaining a strong performance overall the whole space-time domain, especially for longer times, this issue is a matter of research discussed in \ref{subsec_SciML}. 
%
%
%
All errors, MAE, MSE and L2-norm Rel are average over 10 runs.
For all setups, the same optimization parameters are used, including training with $9000$ epochs using Adam with a learning rate of $0.001$ and a final L-BFGS fine-tuning step. 
%
%
%
}
\label{tab:NLS}
\end{table}


\FloatBarrier


\section{PINNs: Data, Applications and Software}

%
The preceding sections cover the neural network component of a PINN framework and which equations have been addressed in the literature.
This section first discusses the physical information aspect, which is how the data and model are managed to be effective in a PINN framework.
Following that, we'll look at some of the real-world applications of PINNs and how various software packages such as DeepXDE, NeuralPDE, NeuroDiffEq, and others were launched in 2019 to assist with PINNs' design. 
%

\subsection{Data}
The PINN technique is based not only on the mathematical description of the problems, embedded in the loss or the NN, but also on the information used to train the model, which takes the form of training points, and impacts the quality of the predictions. Working with PINNs requires an awareness of the challenges to be addressed, i.e. knowledge of the key constitutive equations, and also experience in Neural Network building.

Moreover, the learning process of a PINN can be influenced by the relative magnitudes of the different physical parameters. For example to address this issue, \cite{Kis2020_MachineLearningCardiovascular_YanKYH}  used a non-dimensionalization and normalization technique.

However, taking into account the geometry of the problem can be done without effort.
PINNs do not require fixed meshes or grids, providing greater flexibility in solving high-dimensional problems on complex-geometry domains. 
\\
\noindent
The training points for PINNs can be arbitrarily distributed in the spatio-temporal domain \citep{Pan2019_FpinnsFractionalPhysics_LuPLK}.
However, the distribution of training points influences the flexibility of PINNs. 
Increasing the number of training points obviously improves the approximation, although in some applications, the location of training places is crucial. 
Among the various methods for selecting training points, \cite{Pan2019_FpinnsFractionalPhysics_LuPLK} addressed lattice-like sampling, i.e. equispaced, and quasi-random sequences, such as the Sobol sequences or the Latin hypercube sampling.

Another key property of PINN is its ability to describe latent nonlinear state variables that are not observable.
For instance, \cite{Zha2020_PhysicsInformedMulti_LiuZLS} 
observed that when a variable's measurement was missing, the PINN implementation was capable of accurately predicting that variable.


\subsection{Applications}

In this section, we will explore the real-world applications of PINNs, focusing on the positive leapfrog applications and innovation that PINNs may bring to our daily lives, as well as the implementation of such applications,
such as the ability to collect data in easily accessible locations and simulate dynamics in other parts of the system, or applications to hemodynamics flows, elastic models, or  geoscience.

\paragraph{Hemodynamics}
Three flow examples in hemodynamics applications are presented in
\cite{Sun2020_SurrogateModelingFluid_GaoSGPW}, for addressing either stenotic flow and aneurysmal flow, with standardized vessel geometries and varying viscosity.
In the paper, they not only validate the results against CFD benchmarks, but they also estimate the solutions of the parametric Navier–Stokes equation without labeled data by designing a DNN that enforces the initial and boundary conditions and training the DNN by only minimizing the loss on conservation equations of mass and momentum. 

In order to extract  velocity and pressure fields directly from images, 
\cite{Rai2020_HiddenFluidMechanics_YazRYK}
proposed the  Hidden fluid mechanics (HFM) framework.
The general structure could be applied for electric or magnetic fields in engineering and biology.
They specifically apply it to hemodynamics in a three-dimensional intracranial aneurysm.

Patient-specific systemic artery network topologies can make precise predictions about flow patterns, wall shear stresses, and pulse wave propagation. 
Such typologies of systemic artery networks are used to estimate Windkessel model parameters \citep{Kis2020_MachineLearningCardiovascular_YanKYH}.  
PINN methodology is applied on a simplified form of the  Navier–Stokes equations, where a hyperbolic conservation law defines the evolution of blood velocity and cross-sectional area instead of mass and momentum conservation. 
\cite{Kis2020_MachineLearningCardiovascular_YanKYH} devise a new method for creating 3D simulations of blood flow:
they estimate pressure and retrieve flow information using data from medical imaging.
Pre-trained neural network models can quickly be changed to a new patient condition.
This allows Windkessel parameters to be calculated as a simple post-processing step, resulting in a straightforward approach for calibrating more complex models.
By processing noisy measurements of blood velocity and wall displacement, \cite{Kis2020_MachineLearningCardiovascular_YanKYH}  present a physically valid prediction of flow and pressure wave propagation derived directly from non-invasive MRI flow.
They train neural networks to provide output that is consistent with clinical data.

\paragraph{Flows problems} 
\cite{Mat2021_UncoveringTurbulentPlasma_FraMFH} observe the possibility to infer 3D turbulent fields from only 2D data.
To infer unobserved field dynamics from partial observations of synthetic plasma,
they
simulate the drift-reduced Braginskii model using physics-informed neural networks (PINNs) trained to solve supervised learning tasks while preserving nonlinear partial differential equations.
This paradigm is applicable to the study of quasineutral plasmas in magnetized collisional situations and provides paths for the construction of plasma diagnostics using artificial intelligence.
This methodology has the potential to improve the direct testing of reduced turbulence models in both experiment and simulation in ways previously unattainable with standard analytic methods. 
As a result, this deep learning technique for diagnosing turbulent fields is simply transferrable, allowing for systematic application across magnetic confinement fusion experiments.
The methodology proposed in \cite{Mat2021_UncoveringTurbulentPlasma_FraMFH} can be adapted to many contexts in the interdisciplinary research (both computational and experimental) of magnetized collisional plasmas in propulsion engines and astrophysical environments.

\cite{Xia2020_FlowsOverPeriodic_WuXWLD} examine existing turbulent flow databases and proposes benchmark datasets by methodically changing flow conditions. 


In the context of  high-speed aerodynamic flows, 
\cite{Mao2020_PhysicsInformedNeural_JagMJK},
investigates Euler equations solutions approximated by PINN. 
The authors study both forward and inverse, 1D and 2D, problems.
As for inverse problems, they analyze two sorts of problems that cannot be addressed using conventional approaches. In the first problem they determine the density, velocity, and pressure using data from the density gradient; in the second problem, they determine the value of the parameter in a two-dimensional oblique wave equation of state by providing density, velocity, and pressure data. 

Projecting solutions in time beyond the temporal area used in training is hard to address with the vanilla version of PINN, and such problem is discussed and tested in \cite{Kim2021_DpmNovelTraining_LeeKLL}.
The authors show that vanilla PINN performs poorly on extrapolation tasks in a variety of Burges' equations benchmark problems, and provide a novel NN with a different training approach.

PINN methodology is also used also to address the 1D Buckley-Leverett two-phase problem used in petroleum engineering that has a non-convex flow function with one inflection point, making the problem quite complex \citep{Abr2021_StudyFeedforwardNeural_FloAF}. Results are compared with those obtained by the Lagrangian-Eulerian and Lax-Friedrichs schemes.

Finally, Buckley-Leverett's problem is also addressed in \cite{Alm2022_PredictionPorousMedia_AbuAA}, where PINN is compared to an ANN without physical loss:
when only early-time saturation profiles are provided as data, ANN cannot predict the solution.

\paragraph{Optics and  Electromagnetic applications}
The hybrid PINN from \cite{Fan2021_HighEfficientHybrid_Fan}, based on CNN and local fitting method, addresses applications such as the 3-D Helmholtz equation, quasi-linear PDE operators, and inverse problems.
Additionally, the author tested his hybrid PINN on an icosahedron-based mesh produced by Meshzoo for the purpose of solving surface PDEs. 

A PINN was also developed to address power system applications \citep{Misyris2020PhysicsInformedNN} by solving the swing equation, which has been simplified to an ODE. The authors went on to expand on their research results in a subsequent study \citep{Sti2021_PhysicsInformedNeural_MisSMC}.

In \citep{Kov2022_ConditionalPhysicsInformed_ExlKEK} a whole class of microelectromagnetic problems is addressed with a single PINN model, which learns the solutions of classes of eigenvalue problems, related to the nucleation field associated with defects in magnetic materials.

In \cite{Che2020_PhysicsInformedNeural_LuCLK}, the authors solve inverse scattering problems in photonic metamaterials and nano-optics. They use PINNs to retrieve the effective permittivity characteristics of a number of finite-size scattering systems involving multiple interacting nanostructures and multi-component nanoparticles.  
Approaches for building electromagnetic metamaterials are explored in \cite{Fan2020_DeepPhysicalInformed_ZhaFZ}. 
E.g. cloaking problem is addressed in both \cite{Fan2020_DeepPhysicalInformed_ZhaFZ, Che2020_PhysicsInformedNeural_LuCLK}.
A survey of DL methodologies, including PINNs, applied to nanophotonics may be found in \cite{Wie2021_DeepLearningNano_ArbWAA}.

As a benchmark problem for DeepM\&Mnet, \cite{Cai2021_DeepmmnetInferringElectroconvection_WanCWL} 
choose electroconvection, which depicts electrolyte flow driven by an electric voltage and involves multiphysics, including mass, momentum, cation/anion transport, and electrostatic fields. 

\paragraph{Molecular dynamics and materials related applications} 

Long-range molecular dynamics simulation is addressed with multi-fidelity PINN (MPINN) by estimating nanofluid viscosity over a wide range of sample space using a very small number of  molecular dynamics simulations \cite{Isl2021_ExtractionMaterialProperties_ThaITMH}.
The authors were able to estimate system energy per atom, system pressure, and diffusion coefficients, in particular with the viscosity of argon-copper nanofluid.
%
PINNs can recreate fragile and non-reproducible particles, as demonstrated by \cite{Sti2021_ReconstructionNanoscaleParticles_SchSS}, whose network reconstructs the shape and orientation of silver nano-clusters from single-shot scattering images. 
An encoder-decoder architecture is used to turn 2D images into 3D object space. Following that, the loss score is computed in the scatter space rather than the object space.
The scatter loss is calculated by computing the mean-squared difference error between the network prediction and the target's dispersion pattern.
Finally, a binary loss is applied to the prediction in order to reinforce the physical concept of the problem's binary nature. 
They also discover novel geometric shapes that accurately mimic the experimental scattering data.

A multiscale application is done in \cite{Lin2021_OperatorLearningPredicting_LiLLL, Lin2021_SeamlessMultiscaleOperator_MaxLMLK},
where the authors describe tiny bubbles at both the continuum and atomic levels, the former level using the Rayleigh–Plesset equation and the latter using the dissipative particle dynamics technique.
In this paper, the DeepONet architecture \citep{Lu2021_LearningNonlinearOperators_JinLJP} is demonstrated to be capable of forecasting bubble growth on the fly across spatial and temporal scales that differ by four orders of magnitude.

\paragraph{Geoscience and elastostatic problems}

Based on Föppl–von Kármán (FvK) equations, \cite{Li2021_PhysicsGuidedNeural_BazLBZ}
test their model to four loading cases: in-plane tension with non-uniformly distributed stretching forces, central-hole in-plane tension, deflection out-of-plane, and compression buckling.
Moreover stress distribution and displacement field in solid mechanics problems was addressed by
\cite{Hag2021_SciannKerastensorflowWrapper_JuaHJ}. 

For earthquake hypocenter inversion, \cite{Smi2021_HyposviHypocentreInversion_RosSRAM}
 use Stein variational inference with a PINN trained to solve the Eikonal equation as a forward model, and then test the method against a database of Southern California earthquakes.

\paragraph{Industrial application}
A broad range of applications, particularly in industrial processes, extends the concept of PINN by adapting to circumstances when the whole underlying physical model of the process is unknown.

The process of lubricant deterioration is still unclear, and models in this field have significant inaccuracies; this is why
\cite{Yuc2021_HybridPhysicsInformed_ViaYV}
introduced a hybrid PINN for main bearing fatigue damage accumulation calibrated solely through visual inspections.
They use a combination of PINN and ordinal classifier called discrete ordinal classification (DOrC) approach.
The authors looked at a case study in which 120 wind turbines were inspected once a month for six months and found the model to be accurate and error-free. The grease damage accumulation model was also trained using noisy visual grease inspection data.
Under fairly conservative ranking, researchers can utilize the derived model to optimize regreasing intervals for a particular useful life goal.
\\
\noindent
As for, corrosion-fatigue crack growth and bearing fatigue are examples studied in \cite{Via2021_EstimatingModelInadequacy_NasVNDY}.

For simulating heat transmission inside a channel,  Modulus from NVIDIA \citep{Hen2021_NvidiaSimnetAi_NarHNN}  trains on a parametrized geometry with many design variables. 
They specifically change the fin dimensions of the heat sink (thickness, length, and height) to create a design space for various heat sinks \citep{Modulus2021}.
When compared to traditional solvers, which are limited to single geometry simulations, the PINN framework can accelerate design optimization by parameterizing the geometry.
Their findings make it possible to perform more efficient design space search tasks for complex systems \citep{Cai2021_PhysicsInformedNeural_WanCWW}.




\subsection{Software} 

Several software packages, including DeepXDE \citep{lu2021deepxde}, NVIDIA Modulus (previously SimNet) \citep{Hen2021_NvidiaSimnetAi_NarHNN}, PyDEns  \citep{Kor2019_PydensPythonFramework_KhuKKT}, and NeuroDiffEq  \citep{chen2020neurodiffeq} were released in 2019 to make training PINNs easier and faster. 
The libraries all used feed-forward NNs and the automatic differentiation mechanism to compute analytical derivatives necessary to determine the loss function. 
The way packages deal with boundary conditions, whether as a hard constraint or soft constraint, makes a significant difference.
When boundary conditions are not embedded in the NN but are included in the loss, various losses must be appropriately evaluated.
Multiple loss evaluations and their weighted sum complicate hyper-parameter tuning, justifying the need for such libraries to aid in the design of PINNs. %
\\

More libraries have been built in recent years, and others are being updated on a continuous basis, making this a dynamic field of research and development. 
In this subsection, we will examine each library while also presenting a comprehensive synthesis in
Table~\ref{tab:software}.

\paragraph{DeepXDE} 

DeepXDE \citep{lu2021deepxde} was one of the initial libraries built by one of the vanilla PINN authors. 
This library emphasizes its problem-solving capability, allowing it to combine diverse boundary conditions and solve problems on domains with complex geometries. 
They also present residual-based adaptive refinement (RAR), a strategy for optimizing the distribution of residual points during the training stage that is comparable to FEM refinement approaches. RAR works by adding more points in locations where the PDE residual is larger and continuing to add points until the mean residual is less than a threshold limit. 
DeepXDE also supports complex geometry domains based on the constructive solid geometry  (CSG) technique.
%
This package showcases five applications in its first version in 2019, all solved on scattered points: Poisson Equation across an L-shaped Domain, 2D Burgers Equation, first-order Volterra integrodifferential equation, Inverse Problem for the Lorenz System, and Diffusion-Reaction Systems.

Since the package's initial release, a significant number of other articles have been published that make use of it.  
DeepXDE is used by the authors for:
for inverse scattering \citep{Che2020_PhysicsInformedNeural_LuCLK}, or
 deriving mechanical characteristics from materials with MFNNs \citep{Lu2020_ExtractionMechanicalProperties_DaoLDK}.
 %
While other PINNs are implemented using DeepXDE, like hPINN \citep{Kha2021_HpVpinnsVariational_ZhaKZK}.
Furthermore, more sophisticated tools are built on DeepXDE, such as
DeepONets  \citep{Lu2021_LearningNonlinearOperators_JinLJP}, and its later extension
DeepM\&Mnet \citep{Cai2021_DeepmmnetInferringElectroconvection_WanCWL, Mao2021_DeepmmnetHypersonicsPredicting_LuMLM}.
DeepONets and their derivatives
are considered by the authors to have the significant potential in approximating operators and addressing multi-physics and multi-scale problems, 
like inferring bubble dynamics
\citep{Lin2021_OperatorLearningPredicting_LiLLL, Lin2021_SeamlessMultiscaleOperator_MaxLMLK}.
%

Finally, DeepXDE is utilized for medical ultrasonography applications to simulate a linear wave equation with a single time-dependent sinusoidal source function \citep{Alk2021_ModelingForwardWave_LiuALA},
and the open-source library is also employed for the Buckley-Leverett problem 
\citep{Alm2022_PredictionPorousMedia_AbuAA}.
%
%
A list of research publications that made use of DeepXDE is available online
\footnote{https://deepxde.readthedocs.io/en/latest/user/research.html}
.

\paragraph{NeuroDiffEq} 

NeuroDiffEq \citep{chen2020neurodiffeq} is a PyTorch-based library for solving differential equations with neural networks, which is being used at Harvard IACS. 
NeuroDiffEq solves traditional PDEs (such as the heat equation and the Poisson equation) in 2D by imposing strict constraints, i.e. by fulfilling initial/boundary conditions via NN construction, which makes it a PCNN.
They employ a strategy that is similar to the trial function approach \citep{Lag1998_ArtificialNeuralNetworks_LikLLF}, but with a different form of the trial function. 
However, because NeuroDiffEq enforces explicit boundary constraints rather than adding the corresponding losses, they appear to be inadequate for arbitrary bounds that the library does not support \citep{Bal2021_DistributedMultigridNeural_BotBBK}. 



\paragraph{Modulus} 
Modulus \citep{Modulus2021}, previously known as NVIDIA SimNet  \citep{Hen2021_NvidiaSimnetAi_NarHNN}, from Nvidia, is a toolset for academics and engineers that aims to be both an extendable research platform and a problem solver for real-world and industrial challenges.

It is a  PINN toolbox with support to Multiplicative Filter Networks and a gradient aggregation method for larger batch sizes.
Modulus also offers Constructive Solid Geometry (CSG) and Tessellated Geometry (TG) capabilities, allowing it to parameterize a wide range of geometries. 

In terms of more technical aspects of package implementations, Modulus uses an integral formulation of losses rather than a summation as is typically done.
Furthermore, global learning rate annealing is used to fine-tune the weights parameters $\omega$ in the loss equation~\ref{eq:loss_pinn}.
Unlike many other packages, Modulus appears to be capable of dealing with a wide range of PDE in either strong or weak form.
Additionally, the toolbox supports a wide range of NN, such as Fourier Networks and the DGM architecture, which is similar to the LSTM architecture.

Nvidia showcased the PINN-based code to address multiphysics problems like heat 
transfer in sophisticated parameterized heat sink geometry \citep{Che2021_RecentAdvanceMachine_SeeCS},  3D blood flow in Intracranial Aneurysm or address data assimilation and inverse problems on a flow passing a 2D cylinder \citep{Modulus2021}.
Moreover, Modulus solves the heat transport problem more quickly than previous solvers.

\paragraph{SciANN}  
SciANN \citep{Hag2021_SciannKerastensorflowWrapper_JuaHJ} is an implementation of PINN as a high-level Keras wrapper.
Furthermore, the SciANN repository collects a wide range of examples so that others can replicate the results and use those examples to build other solutions, such as elasticity, structural mechanics, and vibration applications. 
%
%
SciANN is used by the same authors also for  creating a  nonlocal PINN method in \cite{Hag2021_NonlocalPhysicsInformed_BekHBMJ}, or
for a PINN multi-network model applied on solid mechanics \citep{Hag2021_PhysicsInformedDeep_RaiHRM}.
%
Although tests for simulating 2D flood, on Shallow Water Equations, are conducted using SciANN 
\citep{Jam2021_MachineLearningAccelerating_HagJHI}, the authors wrote the feedforward step into a separate function to avoid the overhead associated with using additional libraries. 
%
A general comparison of many types of mesh-free surrogate models based on machine learning (ML) and deep learning (DL) methods is presented in \cite{Hof2021_MeshFreeSurrogate_GeiHGOK}, where SciANN is used among other toolsets.
%
%
%
Finally, the PINN framework for solving the Eikonal equation by \cite{Wah2021_PinneikEikonalSolution_HagWHA} was implemented using SciAnn.
%

\paragraph{PyDENs} 

PyDENs \citep{Kor2019_PydensPythonFramework_KhuKKT} is an open-source neural network PDE solver that allows to define and configure the solution of heat and wave equations.
It impose initial/boundary conditions in the NN, making it a PCNN.
%
After the first release in 2019, the development appears to have stopped in 2020.

\paragraph{NeuralPDE.jl}  

NeuralPDE.jl is part of SciML, a collection of tools for scientific machine learning and differential equation modeling.
In particular SciML (Scientific Machine Learning) \citep{Rac2021_UniversalDifferentialEquations_MaRMM}  is a program written in Julia that combines physical laws and scientific models with machine learning techniques.

\paragraph{ADCME} 

ADCME \citep{Xu2020_AdcmeLearningSpatially_DarXD} can be used to develop numerical techniques and connect them to neural networks: in particular, ADCME was developed by extending and enhancing the functionality of TensorFlow. 
%
In \cite{Xu2020_AdcmeLearningSpatially_DarXD}, ADCME is used to solve different examples, like nonlinear elasticity, Stokes problems, and Burgers' equations.
Furthermore, ADCME is used by \cite{Xu2021_SolvingInverseProblems_DarXD} for solving inverse problems in stochastic models by using a neural network to approximate the unknown distribution.

\paragraph{Nangs} 
Nangs \citep{Ped2019_SolvingPartialDifferential_MarPMP} is a Python library that uses the PDE's independent variables as NN  (inputs), it then computes the derivatives of the dependent variables (outputs), with the derivatives they calculate the PDEs loss function used during the unsupervised training procedure. 
It has been applied and tested on a 1D and 2D advection-diffusion problem.
After a release in 2020, the development appears to have stopped. 
Although, NeuroDiffEq and Nangs libraries were found to outperform PyDEns in solving higher-dimensional PDEs
\citep{Pra2021_AnnsBasedMethod_BakPBMM}.

\paragraph{TensorDiffEq}  
TensorDiffEq \citep{mcclenny2021tensordiffeq} is a Scientific Machine Learning PINN based toolkit on Tensorflow for Multi-Worker Distributed Computing.
Its primary goal is to solve PINNs (inference) and inverse problems (discovery) efficiently through scalability.
It implements a Self-Adaptive PINN to increase the weights when the corresponding loss is greater; this task is accomplished by training the network to simultaneously minimize losses and maximize weights.

\paragraph{IDRLnet}  

IDRLnet \citep{Pen2021_IdrlnetPhysicsInformed_ZhaPZZ} is a Python's PINN toolbox inspired by Nvidia SimNet  \citep{Hen2021_NvidiaSimnetAi_NarHNN}. It provides a  way to mix geometric objects, data sources, artificial neural networks, loss metrics, and optimizers.
It can also solve noisy inverse problems, variational minimization problem, and integral differential equations.

\paragraph{Elvet}  
Elvet \citep{Ara2021_ElvetNeuralNetwork_CriACS} is a Python library for solving differential equations and variational problems.
It can solve systems of coupled ODEs or PDEs (like the quantum harmonic oscillator) and variational problems involving minimization of a given functional (like the catenary or geodesics solutions-based problems).

\paragraph{Other packages} 
Packages that are specifically created for PINN can not only solve problems using PINN, but they can also be used to provide the groundwork for future research on PINN developments.
However, there are other packages that can take advantage of future research developments, such as techniques based on kernel methods \citep{Kar2021_PhysicsInformedMachine_KevKKL}.
Indeed, rather than switching approaches on optimizers, losses, and so on, an alternative approach with respect to PINN framework is to vary the way the function is represented.
Throughout this aspect, rather than using Neural Networks, a kernel method based on Gaussian process could be used. 
The two most noteworthy Gaussian processes toolkit are the Neural Tangents \citep{neuraltangents2020} kernel (NTK), based on JAX, and GPyTorch \citep{gardner2018gpytorch}, written using PyTorch. 
Neural Tangents handles infinite-width neural networks, allowing for the specification of intricate hierarchical neural network topologies. While GPyTorch models Gaussian processes based on Blackbox Matrix-Matrix multiplication using a specific preconditioner to accelerate convergence.

%
%


\afterpage{
\begin{landscape}
\begin{table}[hbt]
\centering
\resizebox{1.67\textwidth}{!}{
\begin{tabular}{@{}  l l l l p{3cm} p{2cm} p{2cm} l @{}}
\toprule
\textbf{Software Name} &
  \textbf{Backend} &
  \textbf{Available for} &
  \textbf{Usage} &
  \textbf{License} &
  \textbf{First release} &
  \textbf{Latest version} &
  \textbf{Code Link} \\ \midrule
DeepXDE \cite{lu2021deepxde} &
  TensorFlow &
  Python &
  Solver &
  Apache-2.0 License &
  v0.1.0  - Jun 13, 2019 &
  v1.4.0 May 20, 2022 &
  \gitlink{lululxvi}{deepxde} \\ \midrule
PyDEns \cite{Kor2019_PydensPythonFramework_KhuKKT} &
  Tensorflow &
  Python &
  Solver &
  Apache-2.0 License &
  v alpha - Jul 14, 2019 &
  v1.0.2 - Jan 20, 2022&
  \gitlink{analysiscenter}{pydens} \\ \midrule
NVIDIA Modulus \cite{Hen2021_NvidiaSimnetAi_NarHNN} &
  TensorFlow &
  Python based API &
  Solver &
  Proprietary &
 v21.06 on Nov 9, 2021 &  %
 v22.03 on Apr 25, 2022 &  %
  \myref{https://developer.nvidia.com/modulus-downloads}{modulus} \\ \midrule
NeuroDiffEq \cite{chen2020neurodiffeq} &
  PyTorch &
  Python &
  Solver &
  MIT License &
  v alpha - Mar 24, 2019 &
  v0.5.2 on Dec 12, 2021 &
  \gitlink{NeuroDiffGym}{neurodiffeq} \\ \midrule
SciANN \cite{Hag2021_SciannKerastensorflowWrapper_JuaHJ} &
  TensorFlow &
  Python 2.7-3.6 &
  Wrapper &
  MIT License &
  v alpha - Jul 21, 2019 &
  v0.6.5.0 Sep 9 21 &
  \gitlink{sciann}{sciann} \\ \midrule
NeuralPDE \cite{Zub2021_NeuralpdeAutomatingPhysics_McCZMM} &
  Julia &
  Julia &
  Solver &
  MIT License   & 
  v0.0.1 - Jun 22, 2019 &
  v4.9.0 - May 26, 2022 &
  \gitlink{SciML}{NeuralPDE.jl} \\ \midrule
ADCME \cite{Xu2020_AdcmeLearningSpatially_DarXD} &
  Julia TensorFlow &
  Julia &
  Wrapper &
  MIT License &
  v alpha - Aug 27, 2019 &
  v0.7.3 May 22, 2021 &
  \gitlink{kailaix}{ADCME.jl} \\ \midrule
Nangs \cite{Ped2019_SolvingPartialDifferential_MarPMP} &
  Pytorch &
  Python &
  Solver &
  Apache-2.0 License &
  v0.0.1 - Jan 9, 2020 &
  v2021.12.6 - Dec 5, 2021 &
  \gitlink{juansensio}{nangs} \\ \midrule
TensorDiffEq \cite{mcclenny2021tensordiffeq} &
  Tensorflow 2.x &
  Python &
  Solver &  
  MIT License &
  v0.1.0 - Feb 03, 2021 &
  v0.2.0 - Nov 17 2021 &
  \gitlink{tensordiffeq}{TensorDiffEq} \\ \midrule
IDRLnet \cite{Pen2021_IdrlnetPhysicsInformed_ZhaPZZ} &
  PyTorch, Sympy &
  Python &
  Solver &  
  Apache-2.0 License &
  v0.0.1-alpha Jul 05, 2021 &
  v0.0.1  Jul 21, 2021 &
  \gitlink{idrl-lab}{idrlnet} \\ \midrule
Elvet \cite{Ara2021_ElvetNeuralNetwork_CriACS} &
  Tensorflow &
  Python &
  Solver & 
  MIT License &
  v0.1.0 Mar 26, 2021 &
  v1.0.0 Mar 29, 2021 &
  \myref{https://gitlab.com/elvet/elvet}{elvet} \\ \midrule
GPyTorch \cite{gardner2018gpytorch} &
  PyTorch &
  Python &
  Wrapper &
  MIT License &
  v0.1.0 alpha Oct 02, 2018 &
  v1.6.0  Dec 04, 2021 &
  \gitlink{cornellius-gp}{gpytorch} \\ \midrule
Neural Tangents \cite{neuraltangents2020} &
  JAX &
  Python &
  Wrapper &
  Apache-2.0 License &
  v0.1.1  Nov 11, 2019 &
  v0.5.0 - Feb 23, 2022 &
  \gitlink{google}{neural-tangents} \\ \bottomrule
\end{tabular}%
}
\caption{Major software libraries specifically designed for physics-informed machine learning}
\label{tab:software}
\end{table}
\end{landscape}
}

\FloatBarrier


%




\section{PINN Future Challenges and directions}

What comes next in the PINN theoretical or applied setup is unknown.
What we can assess here are the paper's incomplete results, which papers assess the most controversial aspects of PINN,  where we see unexplored areas, and where we identify intersections with other disciplines.

Although several articles have been published that have enhanced PINNs capabilities, there are still numerous unresolved issues, such as various applications to real-world situations and equations. 
These span from more theoretical considerations (such as convergence and stability) to implementation issues (boundary conditions management, neural networks design, general PINN architecture design, and optimization aspects).
PINNs and other DL methods using physics prior have the potential to be an effective way of solving high-dimensional PDEs, which are significant in physics, engineering, and finance.
PINNs, on the other hand, struggle to accurately approximate the solution of PDEs when compared to other numerical methods designed for a specific PDE, in particular, they can fail to learn complex physical phenomena, like solutions that exhibit multi-scale, chaotic, or turbulent behavior.

\subsection{Overcoming theoretical difficulties in PINN}

A PINN can be thought of as a three-part modular structure, with an approximation (the neural network),  a module to define how we want to correct the approximation (the physics informed network, i.e. the loss), and the module that manages the minimization of the losses.
The NN architecture defines how well the NN can approximate a function, and the error we make in approximating is called approximation error, as seen in Section~\ref{sec:Theory}. 
Then, how we decide to iteratively improve the approximator will be determined by how we define the loss and how many data points we are integrating or calculating the sum, with the quality of such deviation measured as the generalization error. 
Finally, the quality of iterations in minimizing the losses is dependent on the optimization process, which gives the optimization error.

All of these factors raise numerous questions for future PINN research, the most important of which is whether or not PINN converges to the correct solution of a differential equation. 
The approximation errors must tend to zero to achieve stability, which is influenced by the network topology.
The outcomes of this research are extremely limited. 
For example, the relative error for several neural architectures was calculated by altering the number of hidden layers and the number of neurons per layer in
\cite{Mo2022_DataDrivenVector_LinMLZ}.
In another example, \cite{Ble2021_ThreeWaysSolve_ErnBE} shows the number of successes (i.e. when training loss that is less than a threshold) after ten different runs and for different network topologies (number of layers, neurons, and activation function).
%
Instead, \cite{Mis2021_EstimatesGeneralizationError_MolMM} obtain error estimates and identify possible methods by which PINNs can approximate PDEs. 
It seems that initial hidden layers may be in charge of encoding low-frequency components (fewer points are required to represent low-frequency signals) and the subsequent hidden layers may be in charge of representing higher-frequency components 
\citep{Mar2021_OldNewCan_Mar}.
This could be an extension of the Frequency-principle, F-principle \citep{Zhi2020_FrequencyPrincipleFourier_XuZX9}, according to which DNNs fit target functions from low to high frequencies during training, implying a low-frequency bias in DNNs and explaining why DNNs do not generalize well on randomized datasets. 
For PINN, large-scale features should arise first while small-scale features will require multiple training epochs.

The effects of initialization and loss function on DNN learning, specifically on generalization error, should be investigated. 
Many theoretical results treat loos estimation based on quadrature rules, on points selected randomly and identically distributed.
There are some PINN approaches that propose to select collocations points in specific areas of the space-time domain \citep{Nab2021_EfficientTrainingPhysics_GlaNGM}; this should be investigated as well. 
Finally, dynamic loss weighting for PINN appears to be a promising research direction \citep{Nan2022_DevelopingDigitalTwins_HenNHN}.

Optimization tasks are required to improve the performances of NNs, which also holds for PINN. However, given the physical integration, this new PINN methodology will require additional theoretical foundations on optimization and numerical analysis, and dynamical systems theory. 
According to \citep{Wan2021_UnderstandingMitigatingGradient_TenWTP, Wan2022_WhenWhyPinns_YuWYP}, a key issue is to understand the relationship between PDE stiffness and the impact of algorithms as the gradient descent on the  PINNs.

Another intriguing research topic is why PINN does not suffer from dimensionality.
According to the papers published on PINN, they can scale up easily independently of the size of the problems \citep{Mis2021_PhysicsInformedNeural_MolMM}.
The computational cost of PINN does not increase exponentially as the problem dimensionality increases; this property is common to neural network architecture, and there is no formal explanation for such patterns \citep{De2021_ErrorAnalysisPhysics_MisDRM}. 
\cite{Bau2019_DeepLearningAs_KohBK} recently demonstrated that least-squares estimates based on FNN can avoid the curse of dimensionality in nonparametric regression.
While \cite{Zub2021_NeuralpdeAutomatingPhysics_McCZMMa} demonstrates the capability of the PINN technique with quadrature methods in solving high dimensional problems.

In PINN the process of learning gives rise to a predictor, $u_\theta$, which minimizes the empirical risk (loss).
In machine learning theory, the prediction error can be divided into two components: bias error and variance error.
The bias-variance trade-off appears to contradict the empirical evidence of recent machine-learning systems when neural networks trained to interpolate training data produce near-optimal test results.
It is known that a model should balance underfitting and overfitting, as in the typical U curve, according to the bias-variance trade-off.
However, extremely rich models like neural networks are trained to exactly fit (i.e., interpolate) the data.
\cite{Bel2019_ReconcilingModernMachine_HsuBHMM}, demonstrate the existence of a  double-descent risk curve across a wide range of models and datasets, and they offer a mechanism for its genesis. 
The behavior of PINN in such a framework of Learning Theory for Deep Learning remains to be investigated and could lead to further research questions. 
%
In particular, the function class $\mathcal{H}$ of the hypothesis space in which PINN is optimized might be further examined by specifying such space based on the type of differential equations that it is solving and thus taking into account the physics informed portion of the network. 


In general, PINN can fail to approximate a solution, not due to the lack of expressivity in the NN architecture but due to soft PDE constraint optimization problems \cite{Kri2021_CharacterizingPossibleFailure_GhoKGZ}.

\subsection{Improving implementation aspects in PINN} 
When developing a PINN, the PINN designer must be aware that there may be additional configurations that need to be thoroughly investigated in the literature, as well as various tweaks to consider and good practices that can aid in the development of the PINN, by systematically addressing each of the PINN's three modules.
From neural network architecture options to activation function type.
In terms of loss development, the approaching calculation of the loss integral, as well as the implementation of the physical problem from adimensionalization  or the solution space constrains.
Finally, the best training procedure should be designed.
How to implement these aspects at the most fundamental level, for each physical problem appears to be a matter of ongoing research, giving rise to a diverse range of papers,  which we addressed in this survey, and the missing aspects are summarized in this subsection.


Regarding neural networks architectures, there is still a lack of research for non FFNN types, like CNN and RNN, and what involves their theoretical impact on PINNs \citep{Wan2022_WhenWhyPinns_YuWYP}; moreover, as for the FFNNs
many questions remain, like implementations ones regarding the selection of network size \citep{Hag2021_NonlocalPhysicsInformed_BekHBMJ}.
By looking at Table~\ref{tab:NeuralNetworks}, only a small subset of Neural Networks has been instigated as the network architecture for a PINN, and on a quite restricted set of problems.
Many alternative architectures are proposed in the DL literature \citep{Dar2020_SurveyDeepLearning_KumDKAK}, 
and PINN development can benefit from this wide range of combinations in such research fields. 
%
A possible idea could be to apply Fourier neural operator (FNO) \citep{Wen2022_UFnoanEnhanced_LiWLA}, in order to learn a generalized functional space \citep{Raf2022_DsfaPinnDeep_RafRRC}. 
N-BEATS \citep{Ore2020_NBeatsNeural_CarOCCB}, a deep stack of fully connected layers connected with forward and backward residual links, could be used for  time series based phenomena.
Transformer architecture instead could handle long-range dependencies by modeling global relationships in complex physical problems \citep{Kas2021_PhysicsInformedMachine_MusKMA}.
Finally, sinusoidal representation networks (SIRENs) \cite{Sit2020_ImplicitNeuralRepresentations_MarSMB} that are highly suited for expressing complicated natural signals and their derivatives, could be used in PINN. 
Some preliminary findings are already available \citep{Won2022_LearningSinusoidalSpaces_OoiWOGO, Hua2021_SolvingPartialDifferential_LiuHLS}.
A research line is to study if it is better to increase the breadth or depth of the FFNN to improve PINN outcomes.
Some general studies on DNN make different assertions about whether expanding width has the same benefits as increasing depth.
This may prompt the question of whether there is a minimum depth/width below which a network cannot understand the physics \citep{Tor2020_TheoryTrainingDeep_ViaTRVSA}.
The interoperability of PINN will also play an important role in future research \citep{Rud2022_InterpretableMachineLearning_CheRCC}.
%
%
A greater understanding of PINN's activation function is needed.
\cite{Jag2020_AdaptiveActivationFunctions_KawJKK} show that scalable  activation function may be tuned to maximize network performance, in terms of convergence rate and solution correctness.
%
%
%
%
%
Further research can look into alternative or hybrid methods of differentiating the differential equations. 
To speed up PINN training, the loss function in \cite{Chi2022_CanPinnFast_WonCWO}
is defined using numerical differentiation and automatic differentiation.
The proposed can-PINN, i.e. coupled-automatic–numerical differentiation PINN,  shows to be more sample efficient and more accurate than traditional PINN; because PINN with automatic differentiation can only achieve high accuracy with many collocation points. 
%
While the PINNs training points can be distributed spatially and temporally, making them highly versatile, on the other hand, the position of training locations affects the quality of the results.
%
%
One downside of PINNs is that the boundary conditions must be established during the training stage, which means that if the boundary conditions change, a new network must be created \citep{Wie2021_DeepLearningNano_ArbWAA}. 

%
%
As for the loss, it is important to note that a NN will always focus on minimizing the largest loss terms in the weighted equation, therefore all loss terms must be of the same order of magnitude; increasing emphasis on one component of the loss may affect other parts of it.
There does not appear to be an objective method for determining the weight variables in the loss equation, nor does there appear to be a mechanism to assure that a certain equation can be solved to a predetermined tolerance before training begins; these are topics that still need to be researched \citep{Nan2021_ProgressTowardsSolving_HenNHN}.


It seems there has been a lack of research on optimization tasks for PINNs. Not many solutions appear to be implemented,  apart from standard solutions such as Adam 
and BFGS algorithms \citep{Won2021_CanTransferNeuroevolution_GupWGO}.
The Adam algorithm generates a workflow that can be studied using dynamical systems theory, giving a gradient descent dynamic.
More in detail,  to reduce stiffness in gradient flow dynamics, studying the limiting neural tangent kernel is needed.
Given that there has been great work to solve optimization problems or improve these approaches in machine learning, 
there is still room for improvement in PINNs optimization techniques \citep{Sun2020_SurveyOptimizationMethods_CaoSCZZ}.
The L-BFGS-B is the most common BFGS used in PINNs, and it is now the most critical PINN technology \cite{Mar2021_OldNewCan_Mar}.
\noindent
\\
Moreover,  the impact of learning rate on PINN training behavior has not been fully investigated. 
%
Finally, gradient normalization is another key research topic \citep{Nan2022_DevelopingDigitalTwins_HenNHN}. It is an approach that dynamically assigns weights to different constraints to remove the dominance of any component of the global loss function.



It is necessary to investigate an error estimation for PINN. 
One of the few examples comes from \cite{Hil2022_CertifiedMachineLearning_UngHU}, where
using an ODE, they construct an upper bound on the PINN prediction error.
They suggest adding an additional weighting parameter to the physics-inspired part of the loss function, which allows for balancing the error contribution of the initial condition and the ODE residual. 
Unfortunately, only a toy example is offered in this article, and a detailed analysis of the possibility of providing lower bounds on the error estimator needs still to be addressed, as well as an extension to PDEs.


\subsection{PINN in the SciML framework}\label{subsec_SciML}
PINNs, and SciML in general, hold a lot of potential for applying machine learning to critical scientific and technical challenges. 
However, many questions remain unsolved, particularly if neural networks are to be used as a replacement for traditional numerical methods such as finite difference or finite volume.
In
\cite{Kri2021_CharacterizingPossibleFailure_GhoKGZ}
the authors analyze two basic PDE problems of diffusion and convection and show that PINN can fail to learn the ph problem physics when convection or viscosity coefficients are high.
They found that the PINN loss landscape becomes increasingly complex for large coefficients.
This is partly due to an optimization problem, because of the PINN soft constraint.
However, lower errors are obtained when posing the problem as a sequence-to-sequence learning task instead of solving for the entire space-time at once.
These kinds of challenges must be solved if PINN has to be used beyond basic copy-paste by creating in-depth relations between the scientific problem and the machine learning approaches. 
\noindent
\\

Moreover, unexpected uses of PINN can result from applying this framework to different domains.
PINN has been employed as linear solvers for the Poisson equation \cite{Mar2021_OldNewCan_Mar}, by bringing attention to the prospect of using PINN as linear solvers that are as quick and accurate as other high-performance solvers such as PETSc solvers. 
%
PINNs appear to have some intriguing advantages over more traditional numerical techniques, such as the Finite Element Method (FEM), as explained in \cite{lu2021deepxde}.
Given that PINNs approximate functions and their derivatives nonlinearly, whereas FEM approximates functions linearly, PINNs appear to be well suited for broad use in a wide variety of engineering applications.
However, one of the key disadvantages is the requirement to train the NN, which may take significantly longer time than more extensively used numerical methods.

%
%
%
%

On the other hand, PINNs appear to be useful in a paradigm distinct from that of standard numerical approaches.
PINNs can be deployed in an online-offline fashion, with a single PINN being utilized for rapid evaluations of dynamics in real-time, improving predictions.
Moving from 2D to 3D poses new obstacles for PINN. As training complexity grows in general, there is a requirement for better representation capacity of neural networks, a demand for a larger batch size that can be limited by GPU memory, and an increased training time to convergence \citep{Nan2021_ProgressTowardsSolving_HenNHN}. 
%
Another task is to incorporate PINN into more traditional scientific programs and libraries written in Fortran and C/C++, as well as to integrate PINN solvers into legacy HPC applications \citep{Mar2021_OldNewCan_Mar}.
PINN could also be implemented on Modern HPC Clusters, by using  Horovod \citep{Ser2018_HorovodFastEasy_DelSDB}.
Additionally, when developing the mathematical model that a PINN will solve, the user should be aware of pre-normalizing the problem. At the same time, packages can assist users in dealing with such problems by writing the PDEs in a symbolic form, for example, using SymPy.
\noindent
\\

PINNs have trouble propagating information from the initial condition or boundary condition to unseen areas of the interior or to future times as an iterative solver \citep{Jin2021_NsfnetsNavierStokes_CaiJCLK, Dwi2020_PhysicsInformedExtreme_SriDS}.
This aspect has recently been addressed by \cite{Wan2022_RespectingCausalityIs_SanWSP}
that provided a re-formulation of PINNs loss functions that may explicitly account for physical causation during model training. They assess that PINN training algorithms should be designed to respect how information propagates in accordance with the underlying rules that control the evolution of a given system.
With the new implementation they observe considerable accuracy gains, as well as the possibility to assess the convergence of a PINNs model, and so PINN, can run for the chaotic Lorenz system, the Kuramoto–Sivashinsky equation in the chaotic domain, and the Navier–Stokes equations in the turbulent regime. However there is still research to be done for hybrid/inverse problems, where observational data should be considered as point sources of information, and PDE residuals should be minimized at those points before propagating information outwards. 
Another approach is to use ensemble agreement as to the criterion for incorporating new points in the loss calculated from PDEs \citep{Hai2022_ImprovedTrainingPhysics_IliHI}.
The idea is that in the neighborhood of observed/initial data, all ensemble members converge to the same solution, whereas they may be driven towards different incorrect solutions further away from the observations, especially or large time intervals.
\noindent
\\

PINN can also have a significant impact on our daily lives,
as for the example, from \cite{Yuc2021_HybridPhysicsInformed_ViaYV}, where PINNs are used to anticipate grease maintenance; in the industry 4.0 paradigm, they can assist engineers in simulating materials and constructions or analyzing in real-time buildings structures by embedding elastostatic trained PINNs \citep{Hag2021_PhysicsInformedDeep_RaiHRM, Min2020_SurrogateModelComputational_TroMNTKNRZ}.
PINNs also fail to solve PDEs with high-frequency or multi-scale structure 
\citep{Wan2022_WhenWhyPinns_YuWYP, Wan2021_UnderstandingMitigatingGradient_TenWTP, Fuk2020_LimitationsPhysicsInformed_TchFT}.
The region of attraction of a specific equilibria of a given autonomous dynamical system could also be investigated with PINN \citep{Sch2021_LearningEstimateRegions_HaySH}.

However, to employ PINN in a safety-critical scenario it will still be important to analyze stability and focus on the method's more theoretical component. 
Many application areas still require significant work, such as the cultural heritage sector, the healthcare sector, fluid dynamics, particle physics, and the modeling of general relativity with PINN. 
\noindent
\\
It will be important to develop a PINN methodology for stiff problems, as well as use PINN in digital twin applications such as real-time control, cybersecurity, and machine health monitoring \citep{Nan2021_ProgressTowardsSolving_HenNHN}. 
%
Finally, there is currently a lack of PINNs applications in multi-scale applications, particularly in climate modeling \citep{Irr2021_TowardsNeuralEarth_BoeIBS},
although the PINN methodology has proven capable of addressing its capabilities in numerous applications such as bubble dynamics on multiple scales
\citep{Lin2021_OperatorLearningPredicting_LiLLL, Lin2021_SeamlessMultiscaleOperator_MaxLMLK}.

\subsection{PINN in the AI framework}
PINN could be viewed as a building block in a larger AI framework, or other AI technologies could help to improve the PINN framework. 
\\
\noindent

For more practical applications, PINNs can be used as a tool for engaging deep reinforcement learning (DRL) that combines reinforcement Learning (RL) and deep learning.
RL enables agents to conduct experiments to comprehend their environment better, allowing them to acquire high-level causal links and reasoning about causes and effects
\citep{Aru2017_DeepReinforcementLearning_DeiADBB}.
The main principle of reinforcement learning is to have an agent learn from its surroundings through exploration and by defining a reward \citep{Shr2019_ReviewDeepLearning_MahSM}.
%
%
In the DRL framework, the PINNs can be used as agents.
In this scenario, information from the environment could be directly embedded in the agent using knowledge from actuators, sensors, and the prior-physical law,  like in a  transfer learning paradigm.
\\
\noindent

PINNs can also be viewed as an example of merging deep learning with symbolic artificial intelligence.
The symbolic paradigm is based on the fact that intelligence arises by manipulating abstract models of representations and interactions. 
This approach has the advantage to discover features of a problem using logical inference, but it lacks the easiness of adhering to real-world data, as in DL.
A fulfilling combination of symbolic intelligence and DL would provide the best of both worlds. The model representations could be built up directly from a small amount of data with some priors \citep{Gar2019_ReconcilingDeepLearning_ShaGS}.
In the PINN framework, the physical injection of physical laws could be treated as symbolic intelligence by adding reasoning procedures. 
\\
\noindent

Causal Models are intermediate descriptions that abstract physical models while answering statistical model questions 
\citep{Sch2021_CausalRepresentationLearning_LocSLB}.
Differential equations model allows to forecast a physical system's future behavior, assess the impact of interventions, and predict statistical dependencies between variables.
On the other hand, a statistical model often doesn't refer to dynamic processes, but rather how some variables allow the prediction of others when the experimental conditions remain constant.
In this context, a causal representation learning, merging Machine learning and graphical causality, is a novel research topic, and given the need to model physical phenomena given known data, it may be interesting to investigate whether PINN, can help to determine causality when used to solve hybrid (mixed forward and inverse) problems.

\section{Conclusion}\label{sec_Disc}

This review can be considered an in-depth study of an innovation process over the last four years rather than a simple research survey in the field of PINNs.
%
%
Raissi's first research \citep{Rai2017_PhysicsInformedDeep1_PerRPK,Rai2017_PhysicsInformedDeep2_PerRPK}, which developed the PINN framework, focused on implementing a PINN to solve known physical models.
These innovative papers helped PINN methodology gain traction and justify its original concept even more.
Most of the analyzed studies have attempted to personalize the PINNs by modifying the activation functions, gradient optimization procedures, neural networks, or loss function structures.
A border extension of PINNs original idea brings to use in the physical loss function bare minimum information of the model, without using a typical PDE equation, and on the other side to embed directly in the NN structure the validity of initial or boundary conditions.
Only a few have looked into alternatives to automatic differentiation \citep{Fan2021_HighEfficientHybrid_Fan} or at convergence problems \citep{Wan2021_UnderstandingMitigatingGradient_TenWTP, Wan2022_WhenWhyPinns_YuWYP}.
Finally, a core subset of publications has attempted to take this process to a new meta-level by proposing all-inclusive frameworks for many sub-types of physical problems or multi-physics systems \citep{Cai2021_DeepmmnetInferringElectroconvection_WanCWL}.
The brilliance of the first PINN articles \citep{Rai2017_PhysicsInformedDeep1_PerRPK,Rai2017_PhysicsInformedDeep2_PerRPK} lies in resurrecting the concept of optimizing a problem with a physical constraint by approximating the unknown function with a neural network \citep{Dis1994_NeuralNetworkBased_PhaDP} and then extending this concept to a hybrid data-equation driven approach within modern research. 
Countless studies in previous years have approximated the unknown function using ways different than neural networks, such as the kernel approaches \citep{Owh2019_KernelFlowsLearning_YooOY}, or other approaches that have used PDE functions as constraints in an optimization problem \citep{hinze2008optimization}.

However, PINNs are intrinsically driven by physical information, either from the data point values or the physical equation.
The former can be provided at any point in the domain but is usually only as initial or boundary data.
More importantly, the latter are the collocation points where the NN is forced to obey the physical model equation.



We examined the literature on PINNs in this paper, beginning with the first papers from \cite{Rai2017_PhysicsInformedDeep1_PerRPK,Rai2017_PhysicsInformedDeep2_PerRPK} and continuing with the research on including physical priors on Neural Networks.
This survey looks at PINNs, as a collocation-based method for solving differential questions with Neural networks.
Apart from the vanilla PINN solution we look at most of its variants, like variational PINN (VPINN) as well as in its soft form, with loss including the initial and boundary conditions, and its hard form version with boundary conditions encoded in the Neural network structure.

This survey explains the PINN pipeline, analyzing each building block; first, the neural networks, then the loss construction based on the physical model and feedback mechanism.
Then, an overall analysis of examples of equations in which the PINN methodology has been used, and finally, an insight into PINNs on where they can be concretely applied and the packages available.

%
Finally, we can conclude that
numerous improvements are still possible; most notably, in unsolved theoretical issues. There is still potential for development in training PINNs optimally and extending PINNs to solve multiple equations.
%


\bibliography{revBib}


\end{document}